\documentclass{ieeeaccess}

\usepackage{cite}
\usepackage{amsmath,amssymb,amsfonts}
\usepackage{algorithm}
\usepackage{algorithmic}

\usepackage{graphicx}
\usepackage{textcomp}
\usepackage{booktabs}
\usepackage{colortbl}
\definecolor{tablegray}{gray}{0.95}
\usepackage{multirow}
\usepackage{array}
\usepackage{tabularx}
\usepackage{url}
\usepackage{xspace}
\usepackage{microtype}
\usepackage{balance}

\newcolumntype{L}{>{\raggedright\arraybackslash}X}
\newcolumntype{P}[1]{>{\raggedright\arraybackslash}p{#1}}
\newcolumntype{C}[1]{>{\centering\arraybackslash}p{#1}}
\newcolumntype{Y}{>{\raggedright\arraybackslash}X}
\newcolumntype{Z}{>{\centering\arraybackslash}X}
\newcommand{\Tool}[1]{#1}
\newcommand{\Agent}{M-ArtAgent\xspace}

\usepackage{bm}

\makeatletter
\DeclareFontShape{T1}{formata}{m}{sl}{<->ssub * formata/m/it}{}
\DeclareFontShape{T1}{formata}{b}{sl}{<->ssub * formata/b/n}{}
\def\thetable{\@Roman\c@table}
\makeatother

\begin{document}

\history{}
\doi{}

\title{\Agent: Evidence-Based Multimodal Agent for Implicit Art Influence Discovery}

\author{\uppercase{Hanyi Liu\authorrefmark{1}, Zhonghao Jiu\authorrefmark{2}, Minghao Wang\authorrefmark{3}, Yuhang Xie\authorrefmark{4}, and Heran Yang\authorrefmark{5}}}

\address[1]{China Electronics Technology Group Co., Ltd., Beijing 100043, China (e-mail: liuhanyi@cetc.com.cn)}
\address[2]{School of Information Science and Engineering, Southeast University, Nanjing 211189, China (e-mail: 230228222@seu.edu.cn)}
\address[3]{Department of Chemical and Biological Engineering, Hong Kong University of Science and Technology, Hong Kong SAR, China (e-mail: mwangcx@connect.ust.hk)}
\address[4]{University of California San Diego, La Jolla, CA 92093 USA (e-mail: xieyuhang98@gmail.com)}
\address[5]{Northeastern University, Boston, MA 02115 USA (e-mail: heran.yang@foxmail.com)}


\corresp{Corresponding author: Zhonghao Jiu (e-mail: 230228222@seu.edu.cn).}
\tfootnote{This work did not receive any specific grant from funding agencies in the public, commercial, or not-for-profit sectors.}

\begin{abstract}
Implicit artistic influence, although visually plausible, is often undocumented and thus poses a historically constrained attribution problem: resemblance is necessary but not sufficient evidence. Most prior systems reduce influence discovery to embedding similarity or label-driven graph completion, while recent multimodal large language models (LLMs) remain vulnerable to temporal inconsistency and unverified attributions. This paper introduces \Agent, an evidence-based multimodal agent that reframes implicit influence discovery as probabilistic adjudication. It follows a four-phase protocol consisting of Investigation, Corroboration, Falsification, and Verdict governed by a Reasoning and Acting (ReAct)-style controller that assembles verifiable evidence chains from images and biographies, enforces art-historical axioms, and subjects each hypothesis to adversarial falsification via a prompt-isolated critic. Two theory-grounded operators, \Tool{StyleComparator} for W{\"o}lfflinian formal analysis and \Tool{ConceptRetriever} for ICONCLASS-based iconographic grounding, ensure that intermediate claims are formally auditable. On the balanced WikiArt Influence Benchmark-100 (WIB-100) of 100 artists and 2,000 directed pairs, \Agent achieves 83.7\% positive-class F1, 0.666 Matthews correlation coefficient (MCC), and 0.910 area under the receiver operating characteristic curve (ROC-AUC), with leakage-control and robustness checks confirming that the gains persist when explicit influence phrases are masked. By coupling multimodal perception with domain-constrained falsification, \Agent demonstrates that implicit influence analysis benefits from historically grounded adjudication rather than pattern matching alone.
\end{abstract}

\begin{keywords}
Art influence detection, computational art history, evidence-based adjudication, knowledge graphs, LLM agents, multimodal reasoning, ReAct
\end{keywords}

\titlepgskip=-21pt

\maketitle

\section{Introduction}
\label{sec:introduction}

\PARstart{R}{ecent} advances in digitization and open-access initiatives have turned cultural heritage into large-scale multimodal corpora, advancing both computational art history and visual cultural analytics \cite{manovich2015data,elgammal2018shape}. Vision--language models now support multimodal painting analysis \cite{bin2024gallerygpt}, heritage captioning and metadata tasks \cite{cioni2023diffusion,reshetnikov-marinescu-2025-caption,rei2023metadata,yuan2025culti}, while foundation models \cite{radford2021clip,li2023blip2} are being combined with knowledge-enhanced learning \cite{lymperaiou2024keml} and agentic orchestration \cite{wang2024agents,yao2023react}. Surveys of agentic artificial intelligence (AI) systems further stress that robust domain agents require both agent-side and tool-side adaptation \cite{jiang2025agentic}, and new evaluation suites now probe cultural understanding beyond object recognition \cite{nayak2024culturalvqa}. Together, these advances make influence discovery, one of the central questions in art history, increasingly tractable at scale.

However, the specific task of \emph{implicit influence discovery} remains largely uncharted: determining whether an earlier artist plausibly influenced a later one even when documentary evidence is absent. This question is not reducible to ``find-the-closest-image.'' It is \emph{directional} and \emph{historically constrained}: any credible hypothesis must respect temporal precedence, plausible accessibility, and marker specificity. Seen through a causal lens \cite{pearl2009causality}, portfolio resemblance is an observational correlation confounded by shared movements, period conventions, and convergent evolution. A trustworthy system must therefore adjudicate influence under spatiotemporal constraints and provide an explicit, auditable evidence trail rather than a single similarity score.

\subsection{Related Work}

Existing computational paradigms are limited in this regard because they tend to conflate correlation with historically plausible transmission. Purely visual or metric-learning pipelines \cite{saleh2016influence,elgammal2015creativity,ghildyal2025wpclip,ruta2022stylebabel} operate in a time-unaware embedding space; proximity in that space cannot distinguish direct transmission from shared stylistic ancestry. Knowledge-graph methods \cite{castellano2022artgraph,elvaigh2025gnnboost} support relational reasoning, yet their closed-world reliance on pre-existing labels limits their ability to propose or verify \emph{new} visually grounded hypotheses once structured links are missing. Single-step multimodal large language model (LLM) approaches \cite{bin2024gallerygpt} generate fluent narratives but, without multi-step verification, remain vulnerable to spatiotemporal hallucination and overconfident attributions, failure modes that are particularly damaging in historical scholarship.

This paper therefore argues that implicit influence discovery should be treated as an \emph{evidence-based adjudication} problem, analogous to historical forensics: hypotheses are investigated across modalities, corroborated, actively challenged, and only then accepted with calibrated uncertainty \cite{rudin2019interpretable}. To this end, \Agent is introduced as a multimodal agent that follows a four-phase protocol consisting of Investigation, Corroboration, Falsification, and Verdict and reframes influence detection as probabilistic adjudication. Central to this design is \emph{theory front-loading}: art-historical formalisms are encoded as computational primitives that the agent can query and falsify, instead of being appended as post-hoc narration. Specifically, W{\"o}lfflin's formal oppositions define a low-dimensional style subspace, and ICONCLASS provides a directed acyclic concept topology; both are exposed as evidence operators that a Reasoning and Acting (ReAct) controller invokes alongside a prompt-isolated LLM critic for adversarial falsification.

\subsection{Contributions}

The main contributions of this paper are fourfold, each addressing a specific gap in the existing literature:

\textbf{First (Problem Repositioning)}, whereas prior methods treat influence detection as embedding similarity \cite{saleh2016influence,ghildyal2025wpclip} or graph completion \cite{castellano2022artgraph}, this paper is the first to formulate the problem as \emph{evidence-based probabilistic adjudication}, explicitly requiring that visual correlation be substantiated by temporal feasibility, transmission plausibility, and resistance to adversarial counter-hypotheses before an influence claim is accepted. This repositioning bridges a fundamental gap between pattern-matching approaches and the evidential reasoning standards of art-historical scholarship.

\textbf{Second (Theory-Grounded Formalization)}, existing computational approaches either ignore art-historical theory or invoke it only as post-hoc narration. This paper encodes W{\"o}lfflinian formalism as a mathematically rigorous low-dimensional subspace $\mathcal{W}$ and ICONCLASS iconography as a constrained directed acyclic graph (DAG), making both queryable and falsifiable computational objects rather than decorative labels. No prior system embeds canonical art-historical formalisms as first-class evidence operators within an agentic pipeline.

\textbf{Third (Agentic Adjudication Architecture)}, to the best of the authors' knowledge, \Agent is the first system to combine a ReAct-style multimodal controller with a prompt-isolated LLM critic for adversarial falsification in the cultural heritage domain. The critic mechanism addresses a well-known failure mode of generative models, namely confirmation bias, by systematically generating and evaluating counter-hypotheses such as intermediary transmission, convergent evolution, and common sources.

\textbf{Fourth (Balanced-Benchmark Validation)}, the redesigned WikiArt Influence Benchmark-100 (WIB-100) of 100 artists and 2,000 balanced pairs corrects a systematic evaluation weakness in the field: prior benchmarks were class-imbalanced, rewarding trivial YES-biased predictions. On this balanced benchmark, \Agent achieves \textbf{83.7\% positive-class F1}, \textbf{0.666 Matthews correlation coefficient (MCC)}, and \textbf{0.910 area under the receiver operating characteristic curve (ROC-AUC)}, with leakage-control, module-level, and robustness analyses confirming that the gains are attributable to the adjudication architecture rather than to memorized labels or class-prior exploitation.

In this paper, ``discovery'' denotes a two-stage pipeline: high-recall candidate generation followed by pairwise evidence adjudication. The controlled WIB-100 evaluation focuses on the adjudication stage so that discrimination can be measured without conflating it with open-world retrieval recall.

Section~\ref{sec:preliminaries} introduces the art-historical primitives; Sections~\ref{sec:methodology}--\ref{sec:conclusion} present the methodology, experiments, and conclusion.

\section{Preliminaries: Art-Historical Formalism}

\label{sec:preliminaries}

To ensure mathematical consistency across art-historical theory and the computational architecture, Table~\ref{tab:notation} summarizes the primary notations used in this paper.

\begin{table*}[!t]
\caption{Summary of key mathematical notations.}
\label{tab:notation}
\centering
\small
\setlength{\tabcolsep}{6pt}
\renewcommand{\arraystretch}{1.18}
\begin{tabularx}{\textwidth}{@{}P{0.28\textwidth}Y@{}}
\toprule[1.5pt]
\textbf{Symbol} & \textbf{Meaning} \\
\midrule
$A_i$, $\mathcal{A}_i$, $B_i$ & Artist entity $i$, artwork portfolio, and biographical text corpus \\
$T_i$ & Lifespan interval $[\tau_{\mathrm{birth}}^{(i)},\tau_{\mathrm{death}}^{(i)}]$ \\
$\Delta$ & Formative-window tolerance in the temporal gate \\
$a$ & Artwork instance, with components $a=(I,M)$ \\
$\mathbf{z}(I)$, $\bm{\phi}(I,x)$, $\Omega$ & Global visual embedding, patch-level embedding field, and token index set ($x\in\Omega$) \\
$\alpha_x$, $s_{\mathrm{v}}(\cdot,\cdot)$, $s_{\mathrm{v}}^{\star}$, $\gamma_{\mathrm{v}}$, $K$ & Patch aggregation weights, visual similarity, seed similarity passed from candidate generation, threshold, and top-$K$ retrieval size \\
$\mathcal{W}$, $\mathcal{P}_{\mathcal{W}}$ & W{\"o}lfflinian formal manifold and orthogonal projection operator \\
$\mathbf{w}(I)$, $\bm{\mu}_i$, $D_{\mathcal{W}}(\cdot,\cdot)$ & 5D W{\"o}lfflin coordinate, artist signature, and manifold distance \\
$p_k^{+}$, $p_k^{-}$, $\kappa$ & Textual prototypes defining the two poles of axis $k$ and softmax temperature \\
$\mathcal{G}_{\mathrm{IC}}$, $\mathcal{C}(I)$ & ICONCLASS concept graph and the set of ICONCLASS codes extracted from $I$ \\
$d_{\mathrm{IC}}(\cdot,\cdot)$, $d_{\mathrm{IC}}^{\rightarrow}(\cdot,\cdot)$, $\lambda$ & Topology-aware ICONCLASS distance, directed set distance, and semantic decay factor \\
$\mathcal{T}_{\mathrm{traj}}$ & ReAct execution trajectory $[\mathbf{u}_1,o_1,\ldots]$ \\
$(v,c,\mathcal{E})$, $p_k$, $\omega_k$, $\gamma$ & Verdict tuple, critic plausibility signals, per-hypothesis penalty weights, and global falsification-strength scalar \\
\bottomrule[1.5pt]
\end{tabularx}
\end{table*}

This paper is \emph{theory-grounded} in the sense that the agent is not merely prompted with art-historical terms; instead, canonical art-historical formalisms define explicit computational objects that can be queried, compared, and falsified. Two primitives are introduced that serve as the coordinate system for adjudication: (i) a W{\"o}lfflinian \emph{formal manifold} that captures hereditary stylistic structure beyond raw visual similarity, and (ii) an ICONCLASS \emph{concept topology} that constrains iconographic grounding.

\subsection{W{\"o}lfflinian Formalism as a Latent Formal Manifold}
\label{sec:wolfflin_manifold}

W{\"o}lfflin's five formal oppositions, namely \emph{linear/painterly}, \emph{planar/recessional}, \emph{closed/open form}, \emph{multiplicity/unity}, and \emph{absolute/relative clarity}, can be viewed as a set of (approximately) orthogonal axes describing how visual form is organized \cite{wolfflin1950principles}. These oppositions are formalized in a high-dimensional visual feature space, and a compact, interpretable coordinate representation is derived.

\textbf{High-dimensional visual feature space.} Let $f_{\mathrm{v}}$ be a visual encoder, such as Contrastive Language--Image Pre-training (CLIP), that maps an artwork image $I$ to a global embedding
\begin{equation}
\mathbf{z}(I) = f_{\mathrm{v}}(I) \in \mathbb{R}^{d}.
\end{equation}
To make the definition compatible with both global and local evidence, a patch-level embedding field $\bm{\phi}(I,x)\in\mathbb{R}^{d}$ over spatial locations $x\in\Omega$ is also defined such that $\mathbf{z}(I)$ can be understood as an aggregation of $\bm{\phi}(I,x)$.

\textbf{Orthogonal basis induced by W{\"o}lfflin oppositions.} For each opposition $k\in\{1,\dots,5\}$, two semantic poles are defined via textual prototypes (prompts) $p_k^{+}$ and $p_k^{-}$, such as \emph{painterly} vs.\ \emph{linear}. Let $f_{\mathrm{t}}$ be a text encoder in the same joint space. An (unnormalized) direction is defined as
\begin{equation}
\tilde{\mathbf{b}}_{k} = f_{\mathrm{t}}(p_k^{+}) - f_{\mathrm{t}}(p_k^{-}) \in \mathbb{R}^{d},
\end{equation}
An orthonormal basis $\{\mathbf{b}_{k}\}_{k=1}^{5}$ is then obtained via Gram--Schmidt orthogonalization such that $\mathbf{b}_{k}^{\top}\mathbf{b}_{k'}=\delta_{kk'}$. Let $\mathbf{B}=[\mathbf{b}_1,\dots,\mathbf{b}_5]\in\mathbb{R}^{d\times 5}$.

\textbf{Latent formal manifold and projection.} The W{\"o}lfflinian formal manifold is the 5D subspace $\mathcal{W}=\mathrm{span}(\mathbf{B})$, and the \emph{formal coordinate map} is the orthogonal projection
\begin{equation}
\mathcal{P}_{\mathcal{W}}:\mathbb{R}^{d}\rightarrow\mathbb{R}^{5},\qquad
\mathbf{w}(I)=\mathcal{P}_{\mathcal{W}}(\mathbf{z}(I))=\mathbf{B}^{\top}\mathbf{z}(I).
\label{eq:wolfflin_projection}
\end{equation}
Equivalently, each coordinate admits a discrete aggregation form over local (patch-level) evidence:
\begin{equation}
w_k(I)=\sum_{x\in\Omega}\alpha_x\left\langle \bm{\phi}(I,x),\mathbf{b}_k\right\rangle,
\quad \sum_{x\in\Omega}\alpha_x=1.
\label{eq:wolfflin_sum}
\end{equation}
In practice, $\Omega$ indexes the discrete patch tokens of the visual encoder; in the CLIP instantiation these are Vision Transformer (ViT) tokens. The weights $\alpha_x$ can be uniform ($\alpha_x=1/|\Omega|$) or attention-derived.
The resulting $\mathbf{w}(I)\in\mathbb{R}^{5}$ is a low-dimensional, interpretable descriptor of formal organization. Intuitively, this projection compresses a high-dimensional visual representation into five interpretable scores, each capturing how strongly the artwork leans toward one pole of a W{\"o}lfflinian opposition (e.g., more painterly vs.\ more linear). Two artworks that share an unusual formal profile, such as jointly departing from their period's norm toward recessional depth and open form, thereby produce proximate coordinates in $\mathcal{W}$ even if their surface content differs.

\textbf{Score-based interpretation.} When the joint embedding is contrastive, as in CLIP, each axis also induces a bipolar softmax score:
\begin{equation}
\label{eq:wolfflin_pole_prob}
\begin{aligned}
s_k^{\pm}(I) &= \langle \mathbf{z}(I), f_{\mathrm{t}}(p_k^{\pm})\rangle/\kappa,\\
q_k^{+}(I) &\triangleq \frac{\exp\!\left(s_k^{+}(I)\right)}{\exp\!\left(s_k^{+}(I)\right)+\exp\!\left(s_k^{-}(I)\right)}.
\end{aligned}
\end{equation}

\section{Methodology}
\label{sec:methodology}

Building on the formalism above, this section defines the system model and details the \Agent methodology. Implicit influence verification is cast as a structured reasoning process that goes beyond pattern matching: the system assembles evidence from multiple modalities, tests hypotheses against historical constraints, and returns confidence-scored verdicts for scholarly auditing.

\subsection{System Model and Problem Formulation}
\label{subsec:system_model}

To formalize implicit art influence discovery, artistic transmission is modeled as a heterogeneous multimodal space.

\subsubsection{Problem Scenario}

The overarching scenario involves a large, unannotated repository of digitized artworks and their unstructured biographical corpora. Given any pair of artists in this repository, the system must determine whether a genuine directional stylistic influence exists from the predecessor to the successor. This goes beyond statistical similarity or style classification: the system must infer a historically plausible directional linkage. Artworks serve as visual evidence, while biographies provide the socio-temporal context needed to assess exposure and plausibility.

The central challenge is to filter the vast space of cross-correlations and isolate genuine, transmission-based influence relationships while rejecting those arising from parallel invention, convergent evolution, or shared period styles.

\subsubsection{Raw Inputs and Entities}

The entities processed by the framework are defined as follows.

\textbf{Definition 1 (Artwork Space).} An individual artwork is denoted by $a$, with components $a=(I,M)$. Here, $I\in\mathbb{R}^{H\times W\times 3}$ is the red-green-blue (RGB) image tensor, and $M$ is a metadata tuple capturing basic provenance data: $M=\{m_{\text{artist}},m_{\text{year}},m_{\text{title}},m_{\text{medium}}\}$. Let $\mathcal{A}$ denote the global set of all available artworks. The notation $\operatorname{artist}(a)=m_{\text{artist}}$ is used for the creator identifier extracted from $M$.

\textbf{Definition 2 (Artist Profile).} An artist is modeled as a compound multimodal entity $A_i=(B_i,\mathcal{A}_i,T_i)$, where $B_i$ is the unstructured biographical and historical text corpus associated with artist $i$; $\mathcal{A}_i\subset\mathcal{A}$ is the portfolio authored by artist $i$; and $T_i=[\tau_{\text{birth}}^{(i)},\tau_{\text{death}}^{(i)}]$ denotes the artist's lifespan interval.

\subsubsection{Causal Constraints and the Adjudication Objective}

To declare an implicit influence relationship computationally valid, denoted as a directed edge $R(A_i\rightarrow A_j)$, visual similarity alone is insufficient. The model must enforce structural axioms that mirror established historical scholarship.

\textbf{Definition 3 (Valid Implicit Influence Space).} A directed, probabilistic influence relationship $R(A_i\rightarrow A_j)$ is considered structurally existent if and only if it satisfies three categorical axioms:

\textbf{Temporal Precedence (Chronology):} The putative source must not be later than the target, and it must remain temporally accessible during the target's formative window. Concretely, this requires
    \begin{equation}
    \tau_{\text{birth}}^{(i)} \le \tau_{\text{birth}}^{(j)}
    \quad \text{and} \quad
        \tau_{\text{death}}^{(i)} \ge \tau_{\text{birth}}^{(j)} - \Delta,
    \label{eq:timeline_gate}
    \end{equation}
    where $\Delta=20$ years approximates a conservative early-exposure window. This gate excludes reverse-direction and temporally impossible claims while still allowing overlap between contemporaries and posthumous transmission through durable artifacts.
    
\textbf{Visual and Stylistic Correlation (Specificity):} There must exist a substantive morphological link. Formally, $\exists\,(a_u,a_v)$ such that $a_u\in\mathcal{A}_i$ and $a_v\in\mathcal{A}_j$, where the latent visual similarity $s_{\mathrm{v}}(a_u,a_v)=\cos\big(\mathbf{z}(I_u),\mathbf{z}(I_v)\big)$ exceeds a minimum threshold $\gamma_{\mathrm{v}}$.

\textbf{Evidentiary Transmission (Accessibility):} The textual corpora $B_i$ and $B_j$ must contain socio-historical overlaps, such as geographic co-location, shared terminology, or exhibition intersections, corroborating a plausible transmission pathway.
    
Together, these three axioms function as a structured filter: a candidate influence claim must be temporally possible, visually substantive, and historically transmissible before it can proceed to full adjudication.

\textbf{Definition 4 (Adjudication Mapping).} Influence detection is cast as an evidence-based adjudication mapping rather than a differentiable training loss. Let $\Theta$ denote the fixed parameters of the multimodal reasoning agent. The mapping sends an evaluated artist pair to a verdict tuple:
\begin{equation}
\mathcal{F}_{\Theta}:(A_i,A_j)\mapsto (v,c,\mathcal{E}),
\end{equation}
where $v\in\{\mathrm{YES},\mathrm{NO}\}$ is the terminal verdict, $c\in[0,1]$ is a confidence score, and $\mathcal{E}=\{e_1,e_2,\ldots,e_k\}$ denotes the set of atomic evidence claims supporting $v$.

At inference time, \Agent seeks a verdict whose label is consistent with the multimodal evidence and whose confidence remains stable under adversarial falsification of the supporting evidence chain.

\begin{figure*}[!t]
\centering
\begin{minipage}[b]{0.58\textwidth}
    \centering
    \includegraphics[height=5.5cm]{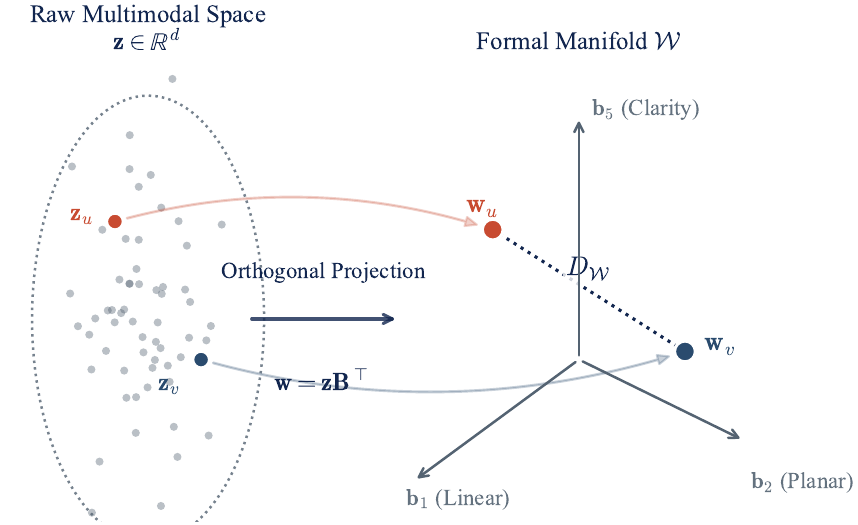}
    \\[1.0ex]
    {\footnotesize\textbf{(a)} The Observation Formalism: W{\"o}lfflin Projection.}
\end{minipage}\hfill
\begin{minipage}[b]{0.38\textwidth}
    \centering
    \includegraphics[height=5.5cm]{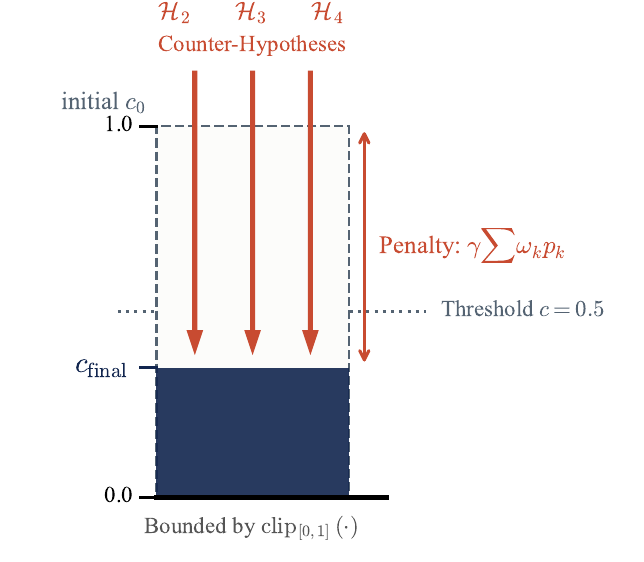}
    \\[1.0ex]
    {\footnotesize\textbf{(b)} The Adjudication Formalism: Critic Clipping.}
\end{minipage}
\caption{Visual summaries of the primary mathematical formalisms in \Agent. \textbf{(a)} The Observation Formalism (Eq. 1--4): Raw multimodal embeddings are funneled through grammatical projection into an orthogonal 5D W{\"o}lfflin manifold $\mathcal{W}$, transforming implicit clustering into explicit formal coordinates. \textbf{(b)} The Adjudication Formalism (Eq.~\ref{eq:confidence_clip}): Rather than blindly accepting visual similarity, the ReAct critic subjects the initial confidence $c_0$ to systematic downward pressure from alternative historical hypotheses ($\mathcal{H}_2, \mathcal{H}_3, \mathcal{H}_4$), actively penalizing and clipping the score to yield a conservative $c_{\mathrm{final}}$.}
\label{fig:formalism}
\end{figure*}
\subsection{System Architecture Overview}

To operationalize this mapping, \Agent integrates four principal processing layers and governs inference with a Four-Phase Adjudication Protocol. Candidate generation is implemented as a high-recall module within Layer 2, not as a standalone intermediate layer. \textbf{Layer~1 (Data Ingestion)} constructs multimodal corpora by aligning high-resolution visual artifacts with corresponding historical texts. \textbf{Layer~2 (Dual-Tower Perception and Candidate Generation)} performs open-vocabulary representation learning via robust visual and textual encoders, builds scalable cross-modal indexes, and applies a high-recall candidate filter under hard historical constraints. \textbf{Layer~3 (Evidence-Based Reasoning, Core)} deploys a ReAct \cite{yao2023react} LLM agent to gather and adjudicate multimodal evidence via specialized tools. \textbf{Layer~4 (Knowledge Graph)} materializes probabilistic verdicts and evidence into a queryable Neo4j graph for macro-level analysis.

\begin{figure*}[!t]
\centering
\begin{minipage}{0.49\textwidth}
    \centering
    \includegraphics[width=\linewidth]{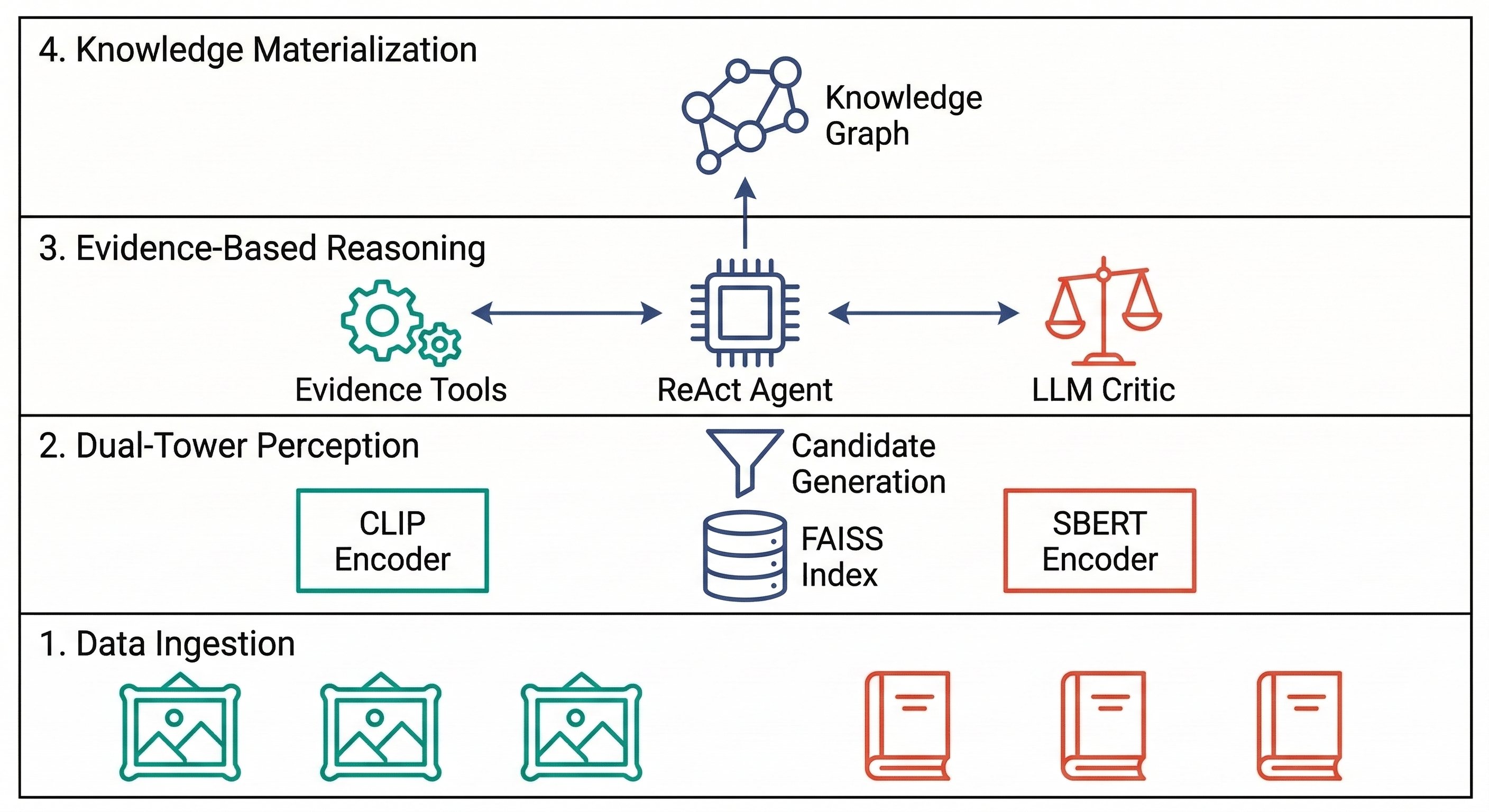}
    \\[-0.5ex]
    {\footnotesize\textbf{(a)} System architecture.}
\end{minipage}\hfill
\begin{minipage}{0.49\textwidth}
    \centering
    \includegraphics[width=\linewidth]{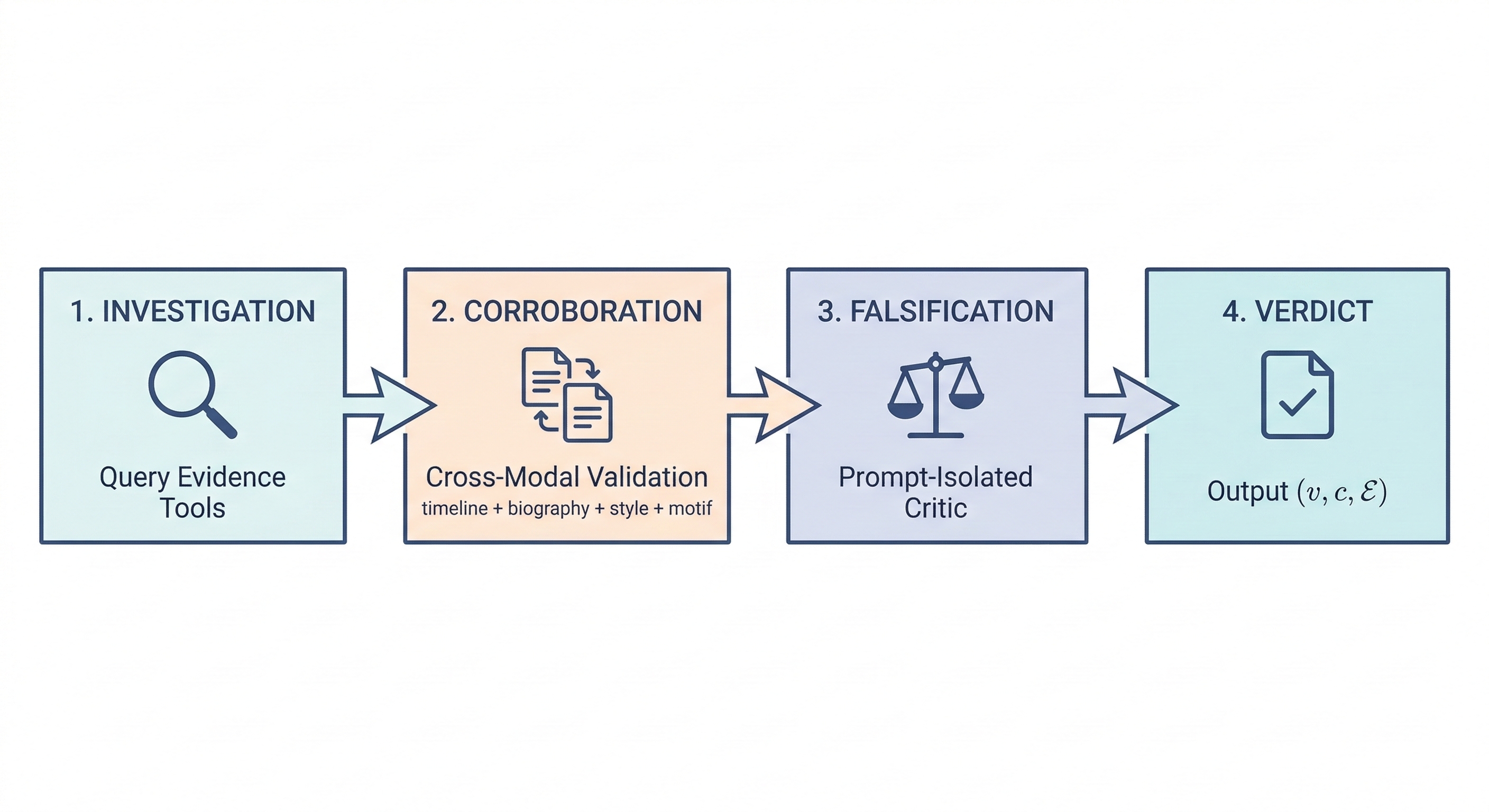}
    \\[-0.5ex]
    {\footnotesize\textbf{(b)} Four-phase adjudication protocol.}
\end{minipage}
\caption{\Agent overview. The framework combines four principal processing layers, with candidate generation integrated into Layer 2, and a four-phase epistemic protocol (Investigation, Corroboration, Falsification, Verdict) that mirrors art-historical scholarship.}
\label{fig:architecture_protocol}
\end{figure*}

\subsubsection{The Four-Phase Adjudication Protocol}

The execution pipeline in Layer 3 follows the stages of formal art-historical research. In the \textbf{Investigation} phase, the agent deploys tools to probe vector spaces and gather heterogeneous evidence. During \textbf{Corroboration}, cross-modal clues are synthesized, resolving contradictions and mutually validating visual similarities with biographical contexts. The \textbf{Falsification} phase then subjects the hypothesis to adversarial stress-testing by a prompt-isolated critic that generates and validates counter-hypotheses. Finally, in the \textbf{Verdict} phase, validated atomic evidence is aggregated to output $(v,c,\mathcal{E})$.

\subsection{Layer 1: Data Ingestion}

The ingestion layer standardizes input modalities into a uniform JavaScript Object Notation (JSON)-based pipeline. For WIB-100, 100 artists are selected from the 129-artist WikiArt pool to improve region and era balance, and each artist portfolio is aligned with a corresponding biographical corpus from Wikipedia. Compared with the original 50-artist benchmark, the share of Impressionist/Post-Impressionist artists drops from 66\% to 38\%, while Renaissance and Baroque coverage rises from 0\% to 23\%.

\subsection{Layer 2: Dual-Tower Perception and Candidate Generation}

To enable open-vocabulary semantic matching and scalable structural comparison, a decoupled dual-encoder perception architecture is deployed.

\subsubsection{Visual Semantic Encoding}

Visual encoding is performed using CLIP \cite{radford2021clip}, initialized with a ViT backbone (ViT-B/32) \cite{dosovitskiy2021vit}. The encoder projects image tensors into an $\ell_2$-normalized subspace representing aesthetic motifs:
\begin{equation}
\mathbf{z}_k = \frac{\mathrm{CLIP}_{\mathrm{visual}}(I_k)}{\|\mathrm{CLIP}_{\mathrm{visual}}(I_k)\|_2} \in \mathbb{R}^{512}.\label{eq:clip_embedding}
\end{equation}

\subsubsection{Textual Biographical Encoding}

Biographical corpora are embedded using Sentence-BERT (SBERT) \cite{reimers2019sbert} (\textit{all-MiniLM-L6-v2}):
\begin{equation}
\mathbf{t}_k = \mathrm{SBERT}(B_k) \in \mathbb{R}^{384}.
\end{equation}

\subsubsection{Approximate Nearest Neighbor Search}
The dual embedding spaces are indexed using Facebook AI Similarity Search (FAISS) \cite{johnson2021faiss} with Hierarchical Navigable Small World (HNSW) graphs \cite{malkov2020hnsw}, enabling efficient sublinear approximate nearest neighbor retrieval in practice. Key implementation details and hyperparameters are summarized in Table~\ref{tab:impl_details}.

\begin{table*}[!t]
\caption{Implementation details and hyperparameters.}
\label{tab:impl_details}
\centering
\footnotesize
\setlength{\tabcolsep}{4pt}
\renewcommand{\arraystretch}{1.12}
\begin{tabularx}{\textwidth}{@{} P{0.18\textwidth} P{0.20\textwidth} X @{}}
\toprule[1.5pt]
\textbf{Section} & \textbf{Component} & \textbf{Configuration} \\
\midrule
\multicolumn{3}{@{}l}{\textbf{Network Architecture: Layer 2 Perception + Candidate Generation}} \\
\midrule
Network Architecture & Visual encoder & CLIP ViT-B/32 \cite{radford2021clip} (512-dim), patch $32\times32$, input $224\times224$ \\
Network Architecture & Text encoder & SBERT \textit{all-MiniLM-L6-v2} \cite{reimers2019sbert} (384-dim), MaxSeq=512, $\ell_2$ normalization \\
Network Architecture & Vector index & FAISS \cite{johnson2021faiss} + HNSW \cite{malkov2020hnsw}, $M=32$, $efConstruction=200$ \\
Network Architecture & Candidate filter & Top-$K=10$, cosine threshold $0.70$, timeline gate (\ref{eq:timeline_gate}); integrated high-recall proposal module within Layer 2 \\
\midrule
\multicolumn{3}{@{}l}{\textbf{Agent Configuration: ReAct + Critic}} \\
\midrule
Agent Configuration & Agent controller & ReAct-style tool use \cite{yao2023react} with five structured tools and a prompt-isolated critic module \\
Agent Configuration & LLM backbone & Claude Opus 4.5, temperature $=0.0$; controller and critic use separate prompts over the same backbone \\
Agent Configuration & Inference hyperparameters & Max ReAct steps $M=8$; default decision threshold $0.5$ before development-fold tuning \\
\midrule
\multicolumn{3}{@{}l}{\textbf{Infrastructure}} \\
\midrule
Infrastructure & Graph database & Neo4j 4.x for probabilistic knowledge graph materialization \\
Infrastructure & Retrieval backend & FAISS-HNSW approximate nearest neighbor search over CLIP/SBERT embeddings \\
\bottomrule[1.5pt]
\end{tabularx}
\end{table*}

\subsubsection{Integrated Candidate-Generation Module}

Exhaustively evaluating all $\mathcal{O}(N^2)$ artist pairs would be prohibitively expensive due to LLM token cost. Within Layer 2, the candidate-generation module acts as a low-latency filter. For each anchor artwork embedding $\mathbf{z}_i$, the top-$K$ nearest neighbors ($K=10$) are retrieved from the visual FAISS index, promoting a candidate pair $(\operatorname{artist}(a_i),\operatorname{artist}(a_j))$ to Layer 3 only if it satisfies three conditions: first, cross-artist uniqueness, i.e., $\operatorname{artist}(a_i)\neq \operatorname{artist}(a_j)$; second, a visual cosine threshold (high-recall), i.e., $s_{\mathrm{v}}(a_i,a_j)\ge \gamma_{\mathrm{v}}$; and third, temporal validity, rendering chronologically feasible mappings under the timeline constraint in \eqref{eq:timeline_gate}.

A permissive similarity threshold ($\gamma_{\mathrm{v}}=0.70$) is intentionally set to maximize recall at the candidate stage, so that visually deceptive hard negatives, such as Dal{\'i}$\rightarrow$Escher with 0.72 similarity, are \emph{kept} and later adjudicated by Layer 3 rather than being prematurely filtered. In the full pipeline, this module therefore behaves as a proposal generator rather than as the final decision component.

For each promoted artist pair, the maximal candidate-stage visual cue is also retained:
\[
s_{\mathrm{v}}^{\star}(A_i,A_j)=\max_{a_u\in\mathcal{A}_i,\;a_v\in\mathcal{A}_j} s_{\mathrm{v}}(a_u,a_v),
\]
which seeds the initial evidence set used by the ReAct controller.

This pruning reduces millions of theoretical pairs to approximately 200--500 high-likelihood candidates for intensive ReAct evaluations. The controlled benchmark in Section~\ref{sec:experiments} evaluates the downstream pair-adjudication stage on curated pairs; separate end-to-end candidate-recall analysis remains future work.

\subsection{Layer 3: Evidence-Based Reasoning (Core)}

\Agent's core is an Evidence-Based Reasoning agent instantiating the ReAct paradigm \cite{yao2023react}.

\subsubsection{Theoretical Forensic State Mapping}

Influence tracking is modeled as an analog of structured forensic procedures: visual trait clues serve as evidence elements, biographies as case files, axiom-based pre-validation as forensic tests, adversarial counter-hypothesis generation as peer scrutiny, and the final ReAct output $(v,c,\mathcal{E})$ as the verdict.

\subsubsection{Specialized Tool Registry}

The reasoning LLM is decoupled from direct web access and interacts only with structured tool interfaces. Each tool returns typed records rather than free-form prose, so both the controller and the critic operate on the same auditable evidence objects.

\begin{table*}[!t]
\caption{Structured evidence operators used by the ReAct controller.}
\label{tab:tool_registry}
\centering
\footnotesize
\setlength{\tabcolsep}{6pt}
\renewcommand{\arraystretch}{1.16}
\begin{tabularx}{\textwidth}{@{} P{0.16\textwidth} P{0.30\textwidth} X @{}}
\toprule[1.5pt]
\textbf{Operator} & \textbf{Evidence produced} & \textbf{Grounding mechanism} \\
\midrule
\Tool{VisualAnalyzer} & Cross-portfolio visual candidates; motif and compositional alignment scores & FAISS-HNSW retrieval in CLIP space together with motif-overlap and spatial-consistency aggregation \\
\Tool{BiographyReader} & Transmission cues such as co-location, institutions, and explicit references & SBERT retrieval with rule-based cue extraction \\
\Tool{TimelineGate} & Chronological feasibility certificate & Rule-based temporal constraints from the chronology axiom \\
\Tool{StyleComparator} & Hereditary formal-profile proximity on $\mathcal{W}$ & Projection onto the W{\"o}lfflin manifold $\mathcal{W}$ using contrastive prompt axes \\
\Tool{ConceptRetriever} & ICONCLASS-aligned motif codes and topology-aware semantic distances & Retrieval and alignment on the ICONCLASS concept graph $\mathcal{G}_{\mathrm{IC}}$ \\
\bottomrule[1.5pt]
\end{tabularx}
\end{table*}

\subsubsection{Theory-Grounded Evidence Operators}
\label{sec:theory_tools}

A central design choice is that art-historical theory is \emph{not} appended as post-hoc interpretation; it defines the computational objects that the agent can query and verify. Concretely, \Tool{StyleComparator} projects artworks onto the W{\"o}lfflinian manifold $\mathcal{W}$ and compares artist-level formal signatures via $D_{\mathcal{W}}$. In parallel, \Tool{ConceptRetriever} maps motifs to ICONCLASS codes and computes topology-aware semantic distances on the concept DAG $\mathcal{G}_{\mathrm{IC}}$. This front-loaded design makes intermediate claims legible through formal coordinates and iconographic codes, so they can be falsified instead of merely narrated post hoc.

In the current implementation, \Tool{ConceptRetriever} converts candidate motif cues into a compact image-level code set $\mathcal{C}(I)$ through graph retrieval followed by hierarchical normalization on $\mathcal{G}_{\mathrm{IC}}$. Both leaf-level codes and their higher-level ancestors are retained so that semantically near matches remain measurable even when exact leaf-code extraction is imperfect. Because this image-to-code interface is crucial to downstream adjudication, Section~\ref{subsec:transparency_checks} reports both exact-code and ancestor-aware alignment quality on a manually checked 100-image sample.

The W{\"o}lfflin manifold is likewise treated as an operational prior rather than as a prompt-invariant truth. In practice, the five axes are induced by fixed pole prompts and orthogonalized once, after which \Tool{StyleComparator} uses the resulting basis throughout all folds. Section~\ref{subsec:transparency_checks} therefore includes axis-order and prompt-perturbation checks to quantify how stable the formal manifold remains under modest design changes.

\subsubsection{Controller Engine and Algorithmic Flow}

The agent interleaves reasoning and acting over a trajectory $\mathcal{T}_{\mathrm{traj}}=[\mathbf{u}_1,o_1,\ldots]$.

\begin{algorithm}[!t]
\caption{ReAct-based probabilistic adjudication flow.}
\label{alg:react_flow}
\begin{algorithmic}[1]
\REQUIRE Candidate artist pair $(A_i, A_j)$, tool registry $\mathcal{T}$, maximum steps $M$
\ENSURE Verdict tuple $(v_{\mathrm{final}}, c_{\mathrm{final}}, \mathcal{E})$

\STATE \COMMENT{Phase 0: axiomatic temporal gate}
\IF{$(\tau_{\mathrm{birth}}^{(i)} > \tau_{\mathrm{birth}}^{(j)}) \vee (\tau_{\mathrm{death}}^{(i)} < \tau_{\mathrm{birth}}^{(j)} - \Delta)$}
    \RETURN $(\mathrm{NO}, 0.95, \{\text{Timeline impossible}\})$
\ENDIF

\STATE \COMMENT{Phase 1: context initialization}
\STATE $\mathcal{E} \leftarrow \{\text{metadata}, s_{\mathrm{v}}^{\star}(A_i,A_j)\}$
\STATE $(v_0, c_0) \leftarrow (\mathrm{NO}, 0.50)$ \COMMENT{conservative fallback}

\STATE \COMMENT{Phase 2: ReAct loop}
\FOR{$s = 1$ \TO $M$}
    \STATE $\mathbf{u}_s \leftarrow \mathrm{LLM}(\mathcal{E})$
    \IF{$\mathbf{u}_s = \mathrm{Call}(t,\mathrm{args})$ for some $t \in \mathcal{T}$}
        \STATE $o_s \leftarrow t(\mathrm{args})$
        \STATE $\mathcal{E} \leftarrow \mathcal{E} \cup \{o_s\}$
    \ELSIF{$\mathbf{u}_s = \mathrm{Conclude}(v', c')$}
        \STATE $(v_0, c_0) \leftarrow (v', c')$
        \STATE break
    \ENDIF
\ENDFOR

\STATE \COMMENT{Phase 3: adversarial falsification}
\STATE $(v_{\mathrm{final}}, c_{\mathrm{final}}) \leftarrow \mathrm{Critic}(v_0, c_0, \mathcal{E})$ \COMMENT{confidence clipping in (\ref{eq:confidence_clip})}

\STATE \COMMENT{Phase 4: verdict rendering}
\RETURN $(v_{\mathrm{final}}, c_{\mathrm{final}}, \mathcal{E})$
\end{algorithmic}
\end{algorithm}

\subsubsection{Prompt-Isolated LLM Critic: Adversarial Falsification}

To mitigate confirmation bias intrinsic to generative models, a prompt-isolated critic role is used, instantiated from the same temperature-controlled backbone but with separate instructions and access only to the provisional verdict and structured evidence set. Given preliminary $(v_0,c_0)$, the critic evaluates three competing counter-hypotheses: $\mathcal{H}_2$ (intermediary), which asks whether influence could flow through a mediator $Z$ ($A_i\rightarrow Z\rightarrow A_j$); $\mathcal{H}_3$ (convergence), which asks whether $A_i$ and $A_j$ could converge independently absent interaction; and $\mathcal{H}_4$ (common source), which asks whether both could share a prior common influencer $Z$.

For each $\mathcal{H}_k$, the critic outputs a plausibility signal $p_k$ and applies a structured confidence penalty:
\begin{equation}
 c_{\mathrm{final}} = \operatorname{clip}_{[0,1]}\!\left(c_0 - \gamma \sum_{k=2}^{4} \omega_k\,p_k \right),
\label{eq:confidence_clip}
\end{equation}
where $\operatorname{clip}_{[0,1]}(x)=\min(1,\max(0,x))$ constrains the score range, $p_k\in[0,1]$ is interpreted as a critic-assigned relative plausibility signal rather than as a calibrated posterior probability, $\omega_k\ge 0$ are fixed per-hypothesis penalty weights, and $\gamma\ge 0$ is a global falsification-strength scalar tuned in the ablation study. In effect, the critic acts as a skeptical second opinion: the more plausible an alternative explanation, such as convergent artistic evolution rather than direct transmission, the more the confidence in direct influence is reduced. Because the controller and critic share the same backbone, the mechanism amounts to adversarial role separation over the same evidence objects rather than an independent second model family.

\subsection{Embedded Domain Axioms and Historical Gate Constraints}

\Agent translates art-historical epistemology into explicit computational constraints. Some act as hard gates, such as temporal impossibility; others act as evidence-sensitive penalties that down-weight historically weak hypotheses rather than forcing an immediate rejection.

\begin{table*}[!t]
\caption{Historical constraints that bound the generative agent.}
\label{tab:axioms}
\centering
\footnotesize
\setlength{\tabcolsep}{6pt}
\renewcommand{\arraystretch}{1.18}
\begin{tabularx}{\textwidth}{@{} P{0.19\textwidth} P{0.19\textwidth} P{0.30\textwidth} P{0.26\textwidth} @{}}
\toprule[1.5pt]
\textbf{Axiom rule} & \textbf{Methodological source} & \textbf{Computational implementation} & \textbf{Consequence} \\
\midrule
Chronological intersection & Biographical chronology & \Tool{TimelineGate} & Immediate rejection, forced negative \\
Pathway accessibility & Social-historical context & SBERT overlap plus transmission cues & Confidence reduction if absent \\
Marker specificity & Connoisseurship-style diagnosis & Style taxonomy plus motif checks & Require diagnostic markers \\
Convergence risk & Comparative stylistic analysis & Critic hypothesis $\mathcal{H}_3$ & Penalize spurious similarity \\
\bottomrule[1.5pt]
\end{tabularx}
\end{table*}

\subsection{Layer 4: Knowledge Graph Generation}

After adjudicating candidate pairs, local evidence and confidence values are serialized into a Neo4j knowledge graph. Nodes include Artist and Artwork; edges encode INFLUENCED relationships with attributes such as \emph{confidence}, \emph{evidence}, and the supporting tool outputs. This graph serves as an interpretable analysis artifact, not as a training signal, allowing accepted and rejected hypotheses to be queried together with their provenance.

\subsection{Implementation Summary}

Implementation details are summarized in Table~\ref{tab:impl_details}. In brief, the framework uses a temperature-controlled Claude Opus 4.5 reasoning backbone at temperature $0.0$, CLIP ViT-B/32 for visual encoding, SBERT for textual encoding, FAISS-HNSW for approximate retrieval, and Neo4j for knowledge graph materialization. The controller and critic are implemented as prompt-isolated roles over the same backbone so that adversarial checking changes the evidence interpretation rather than the underlying model family. Here, ``structured'' refers to schema-constrained evidence access and rule-based gates rather than bitwise reproducibility of approximate nearest neighbor retrieval or LLM internals.


\section{Experiments}
\label{sec:experiments}

This section evaluates \Agent on WIB-100, a redesigned benchmark that directly addresses the class-prior and coverage issues observed in the earlier benchmark configuration. Because WIB-100 is a controlled benchmark of labeled artist pairs, the reported numbers measure pair-level adjudication performance rather than end-to-end open-world retrieval recall. The updated evaluation addresses five research questions (RQs): \textbf{RQ1}, how does \Agent compare with representative baselines on the balanced WIB-100 benchmark; \textbf{RQ2}, how much do critic falsification strength and theory-grounded formal connectors contribute to performance; \textbf{RQ3}, how well does the system reject hard, medium, easy, and temporal-impossible negatives; \textbf{RQ4}, does the model preserve strong threshold-independent ranking performance; and \textbf{RQ5}, do qualitative case studies support the evidence chains returned by the agent.

\subsection{Experimental Setup}
\subsubsection{Dataset}

The benchmark is redesigned as WIB-100, built around 100 artists selected from the 129-artist WikiArt pool. The selection explicitly balances region, era, and movement: the original 50 artists are retained for backward compatibility, while 50 new artists are added to expand Renaissance, Baroque, British, American, Eastern European, Japanese, and Latin American coverage. Relative to the original 50-artist benchmark, the share of Impressionist/Post-Impressionist artists drops from 66\% to 38\%, while Renaissance and Baroque coverage rises from 0\% to 23\%.

\begin{table}[!t]
\caption{Summary of WIB-100.}
\label{tab:dataset_wib100}
\centering
\footnotesize
\setlength{\tabcolsep}{4pt}
\renewcommand{\arraystretch}{1.14}
\begin{tabularx}{\columnwidth}{@{} P{0.25\columnwidth} X @{}}
\toprule[1.5pt]
\textbf{Item} & \textbf{WIB-100 design} \\
\midrule
Artist pool & 100 artists selected from 129 WikiArt candidates; region/era balanced and backward compatible with the original 50 artists \\
Evaluation pairs & 2,000 directed artist pairs \\
Class balance & 1,000 positives + 1,000 negatives (50:50) \\
Negative tiers & 300 hard, 300 medium, 200 easy, and 200 temporal-impossible pairs \\
Protocol & 5-fold stratified cross-validation; each round uses 400 pairs for development and 1,600 for held-out evaluation \\
Ground truth & Positives cross-validated from art-historical literature and curated influence records; negatives actively verified rather than treated as unlabeled pairs \\
\bottomrule[1.5pt]
\end{tabularx}
\end{table}

Positive relations are collected from art-historical literature and curated influence records, and each retained positive relation is cross-validated against multiple sources when available. Rather than sampling negatives from unlabeled pairs, WIB-100 explicitly constructs verified hard, medium, easy, and temporal-impossible negatives. Ambiguous unlabeled pairs are not treated as negatives, which reduces false-negative contamination and makes rejection ability measurable. To avoid prior-driven evaluation and make rejection ability directly measurable, WIB-100 enforces a strict 50:50 positive/negative split.

\subsubsection{Evaluation Metrics on the Balanced Benchmark}
\label{subsec:metrics}

Because WIB-100 is explicitly balanced, positive-class F1 remains comparable to prior work but no longer rewards trivial YES bias. The evaluation therefore reports Precision, Recall, Specificity, Balanced Accuracy, positive-class F1, macro-averaged F1 (Macro-F1), MCC \cite{chicco2020mcc}, and ROC-AUC. Throughout the paper, rate-based metrics are reported in percentages, whereas MCC and ROC-AUC are reported as unitless coefficients; this convention is retained consistently in the text and tables.

Let true positives (TP), false positives (FP), true negatives (TN), and false negatives (FN) denote the confusion-matrix counts. Balanced accuracy and MCC are defined as
\begin{equation}
\begin{aligned}
\mathrm{BalAcc} &= \frac{1}{2}\big(\mathrm{Recall}+\mathrm{Specificity}\big),\\
\mathrm{MCC} &= \frac{\mathrm{TP}\cdot\mathrm{TN}-\mathrm{FP}\cdot\mathrm{FN}}
{\sqrt{(\mathrm{TP}+\mathrm{FP})(\mathrm{TP}+\mathrm{FN})(\mathrm{TN}+\mathrm{FP})(\mathrm{TN}+\mathrm{FN})}}.
\end{aligned}
\end{equation}
On this 50:50 benchmark, an Always-YES classifier yields Precision $=50.0\%$, Recall $=100\%$, Specificity $=0\%$, Balanced Accuracy $=50.0\%$, F1$_{\mathrm{pos}}=66.7\%$, Macro-F1 $=33.3\%$, and MCC $=0.0$ by convention, because the MCC denominator is zero for this degenerate predictor. This gives a meaningful lower bound: methods that cannot surpass it are not genuinely discriminative.

\subsubsection{Baselines}

\Agent{} is compared against nine representative baselines spanning multimodal LLMs, text-only LLMs, CLIP-based models, metric learning, and knowledge-graph (KG) embeddings, together with the Always-YES baseline that exposes residual class-prior bias. To address the concern that graph completion is a relevant comparator, a translation-based KG baseline (TransE) \cite{bordes2013translating} and a stronger complex-valued KG baseline (ComplEx) \cite{trouillon2016complex} are additionally included.

\begin{table*}[!t]
\caption{Representative baselines used in the WIB-100 benchmark. Type abbreviation: multimodal large language model (MLLM).}
\label{tab:baselines}
\centering
\footnotesize
\setlength{\tabcolsep}{4pt}
\renewcommand{\arraystretch}{1.12}
\begin{tabularx}{\textwidth}{@{} P{0.24\textwidth} C{0.06\textwidth} C{0.08\textwidth} X @{}}
\toprule[1.5pt]
\textbf{Method} & \textbf{Year} & \textbf{Type} & \textbf{Description} \\
\midrule
GalleryGPT \cite{bin2024gallerygpt} & 2024 & MLLM & Art-specialized multimodal large language model \\
ComplEx (Text + Vis) \cite{trouillon2016complex} & 2016 & KG & Complex-valued knowledge-graph embedding baseline augmented with multimodal artist descriptors \\
LLM Zero-Shot \cite{kojima2022zerocot} & 2022 & LLM & Zero-shot reasoning protocol instantiated on a GPT-4-class backbone without tailored prompting \\
LLM + Simple CoT \cite{wei2022cot} & 2022 & LLM & Chain-of-thought (CoT) prompting protocol instantiated on a GPT-4-class backbone \\
WP-CLIP \cite{ghildyal2025wpclip} & 2025 & CLIP & W{\"o}lfflin--Panofsky aligned CLIP fine-tuning \\
TransE (DeepWalk) \cite{bordes2013translating} & 2013 & KG & Translation-based knowledge-graph embedding baseline with graph-walk artist features \\
CLIP-Art \cite{conde2021clipart} & 2021 & CLIP & Artistic adaptation of CLIP for art classification \\
Siamese Art \cite{li2023siamese} & 2023 & Metric & Siamese network similarity model \\
Always-YES Baseline & N/A & N/A & Predicts YES for every pair; included to expose whether a method merely exploits class priors \\
\bottomrule[1.5pt]
\end{tabularx}
\end{table*}

\subsubsection{Implementation}

Implementation details and hyperparameters are summarized in Table~\ref{tab:impl_details}. Unless otherwise stated, all experiments use a temperature-controlled LLM backbone at temperature $0.0$, and results are reported as five-fold mean $\pm$ standard deviation. The verdict threshold is initialized at $0.5$, selected on the development fold in each cross-validation round, and then held fixed on the corresponding evaluation split.

\subsubsection{Leakage-Control and Robustness Protocols}

To test whether \Tool{BiographyReader} is merely reading the answer from biographies, a masked-biography setting is constructed in which direct influence predicates are redacted before SBERT retrieval and cue extraction, while the remaining temporal, institutional, and geographic context is preserved. Text-only and no-\Tool{BiographyReader} variants are additionally reported to separate accessibility evidence from visual evidence.

To audit the two theory-grounded operators directly, two lightweight transparency checks are conducted. First, \Tool{ConceptRetriever} is evaluated on a manually checked 100-image sample, reporting both exact leaf-code agreement and ancestor-aware agreement, because topology-aware retrieval is designed to tolerate semantically near misses. Second, the W{\"o}lfflin manifold is tested under axis-order permutation, prompt paraphrasing, and an alternative CLIP embedding to quantify how sensitive the formal connector is to prompt design and encoder choice.

\subsection{Main Results (RQ1)}

\begin{figure*}[!t]
\centering
\includegraphics[width=0.98\textwidth]{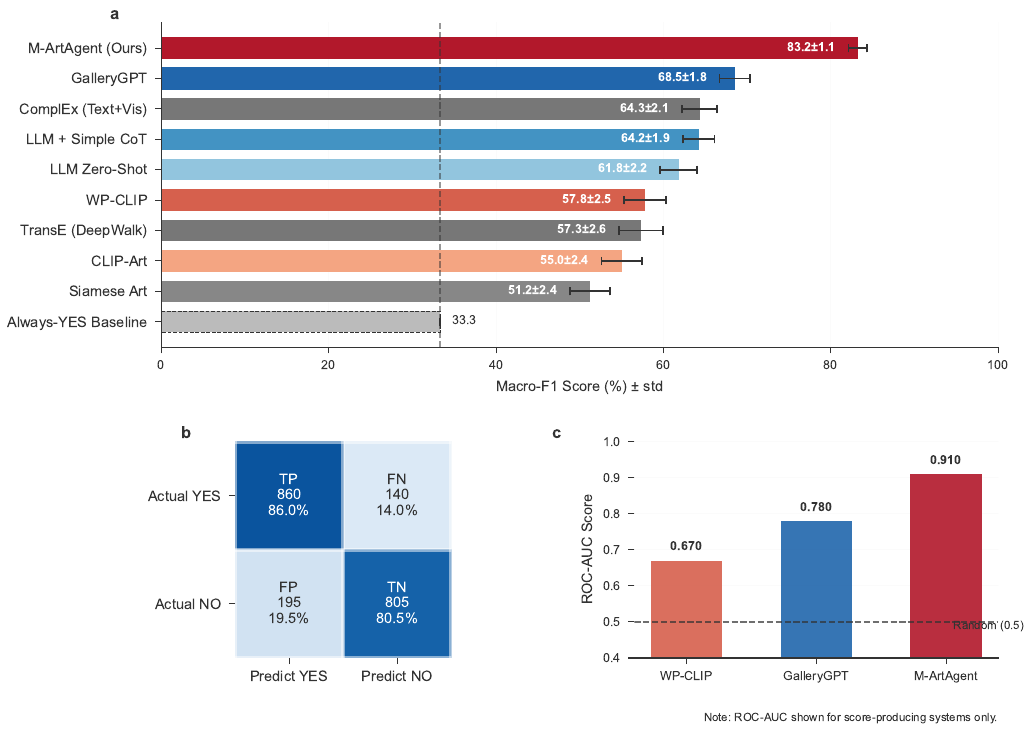}
\caption{Main results visualization on the balanced WIB-100 benchmark for selected baselines. (a) Positive-class F1 comparison; the dashed line marks the Always-YES baseline. (b) Confusion matrix of \Agent on the full 2,000-pair benchmark. (c) ROC-AUC comparison for score-producing systems.}
\label{fig:main_performance}
\end{figure*}

Table~\ref{tab:main_results} reports the main results on the balanced WIB-100 benchmark as five-fold mean $\pm$ standard deviation. A visual overview of selected baselines is provided in Fig.~\ref{fig:main_performance}.

\begin{table*}[!t]
\caption{Main performance comparison on the balanced WIB-100 benchmark.}
\label{tab:main_results}
\centering
\footnotesize
\setlength{\tabcolsep}{4pt}
\renewcommand{\arraystretch}{1.12}
\begin{tabularx}{\textwidth}{@{} P{0.18\textwidth} C{0.08\textwidth} *{6}{Z} @{}}
\toprule[1.5pt]
\textbf{Method} & \textbf{Type} & \textbf{Prec.} & \textbf{Rec.} & \textbf{Spec.} & \textbf{F1$_{\mathrm{pos}}$} & \textbf{Mac-F1} & \textbf{MCC} \\
\midrule
\rowcolor{tablegray}
\textbf{\Agent} & \textbf{Agent} & \textbf{81.5 $\pm$ 1.5\%} & \textbf{86.0 $\pm$ 1.4\%} & \textbf{80.5 $\pm$ 1.8\%} & \textbf{83.7 $\pm$ 1.2\%} & \textbf{83.2 $\pm$ 1.1\%} & \textbf{0.666 $\pm$ 0.021} \\
GalleryGPT \cite{bin2024gallerygpt} & MLLM & 65.1 $\pm$ 2.2\% & 82.0 $\pm$ 2.1\% & 56.0 $\pm$ 3.4\% & 72.6 $\pm$ 2.0\% & 68.5 $\pm$ 1.8\% & 0.394 $\pm$ 0.032 \\
ComplEx (Text+Vis) \cite{trouillon2016complex} & KG & 62.1 $\pm$ 2.6\% & 75.5 $\pm$ 3.4\% & 54.0 $\pm$ 4.1\% & 68.2 $\pm$ 2.3\% & 64.3 $\pm$ 2.1\% & 0.302 $\pm$ 0.039 \\
LLM + Simple CoT \cite{wei2022cot} & LLM & 61.5 $\pm$ 2.4\% & 80.0 $\pm$ 2.5\% & 50.0 $\pm$ 4.1\% & 69.6 $\pm$ 2.2\% & 64.2 $\pm$ 1.9\% & 0.314 $\pm$ 0.038 \\
LLM Zero-Shot \cite{kojima2022zerocot} & LLM & 59.5 $\pm$ 2.8\% & 88.0 $\pm$ 2.8\% & 40.0 $\pm$ 4.5\% & 71.0 $\pm$ 2.5\% & 61.8 $\pm$ 2.2\% & 0.319 $\pm$ 0.045 \\
WP-CLIP \cite{ghildyal2025wpclip} & CLIP & 56.8 $\pm$ 3.5\% & 83.0 $\pm$ 3.0\% & 37.0 $\pm$ 4.2\% & 67.5 $\pm$ 2.6\% & 57.8 $\pm$ 2.5\% & 0.225 $\pm$ 0.040 \\
TransE (DeepWalk) \cite{bordes2013translating} & KG & 56.3 $\pm$ 3.1\% & 71.0 $\pm$ 4.5\% & 45.0 $\pm$ 5.2\% & 62.8 $\pm$ 2.8\% & 57.3 $\pm$ 2.6\% & 0.166 $\pm$ 0.048 \\
CLIP-Art \cite{conde2021clipart} & CLIP & 55.1 $\pm$ 3.0\% & 81.0 $\pm$ 3.2\% & 34.0 $\pm$ 3.8\% & 65.6 $\pm$ 2.7\%$^{\dagger}$ & 55.0 $\pm$ 2.4\% & 0.170 $\pm$ 0.041 \\
Siamese Art \cite{li2023siamese} & Metric & 52.4 $\pm$ 3.4\% & 75.0 $\pm$ 3.5\% & 32.0 $\pm$ 4.0\% & 61.7 $\pm$ 2.6\%$^{\dagger}$ & 51.2 $\pm$ 2.4\% & 0.078 $\pm$ 0.035 \\
\textit{Always-YES Baseline} & --- & 50.0 $\pm$ 0.0\% & 100.0 $\pm$ 0.0\% & 0.0 $\pm$ 0.0\% & 66.7 $\pm$ 0.0\% & 33.3 $\pm$ 0.0\% & 0.000 $\pm$ 0.000 \\
\bottomrule[1.5pt]
\multicolumn{8}{@{}p{\textwidth}@{}}{\vspace{2pt}\scriptsize \textit{Note:} Evaluated on 2,000 artist pairs (50:50 split). Results are reported as 5-fold mean $\pm$ standard deviation. Rate-based metrics are percentages, whereas MCC is unitless. $^{\dagger}$ marks methods falling below the naive Always-YES F1$_{\mathrm{pos}}$ baseline.}
\end{tabularx}
\end{table*}

Table~\ref{tab:main_results} shows that \Agent{} achieves the strongest overall performance among all compared systems. Over five folds, it reaches $83.2\pm1.1\%$ Macro-F1 and $0.666\pm0.021$ MCC while preserving both high recall ($86.0\pm1.4\%$) and high specificity (80.5 $\pm$ 1.8\%). GalleryGPT remains the strongest overall baseline, but it trails by 24.5 points in specificity and by $0.272$ MCC. Among the newly added KG comparators, ComplEx is the stronger baseline at 64.3 $\pm$ 2.1\% Macro-F1 and 0.302 $\pm$ 0.039 MCC, showing that artist-level graph structure helps but still falls well short of pairwise evidence adjudication.

The redesigned benchmark also changes the interpretation of positive-class F1. Because the class prior is now balanced, a model can no longer score well by predicting YES for most pairs. The Always-YES baseline collapses to 66.7 $\pm$ 0.0\% F1$_{\mathrm{pos}}$ and 33.3 $\pm$ 0.0\% Macro-F1, and both CLIP-Art and Siamese Art fall below that naive threshold. This confirms the need for the WIB-100 redesign: the new benchmark rewards selective classification over permissive YES bias.

\subsubsection{Confusion Matrix}

At the operating threshold selected during cross-validation, the aggregated confusion matrix on the full 2,000-pair benchmark is TP$=860$, FN$=140$, FP$=195$, and TN$=805$ (Fig.~\ref{fig:main_performance}\,(b)). In other words, \Agent correctly retrieves 860 of the 1,000 positives while rejecting 805 of the 1,000 negatives, a regime that was largely obscured in the earlier imbalanced benchmark.

\subsubsection{Threshold-Independent Performance (RQ4)}

The confidence scores also remain well ordered across thresholds. As shown in Fig.~\ref{fig:main_performance}\,(c), \Agent reaches a ROC-AUC of 0.910, compared with 0.780 for GalleryGPT and 0.670 for WP-CLIP. This indicates that the gain is not tied to a single operating point: the critic-adjusted confidence produces a materially better ranking of plausible and implausible influence pairs.

\subsection{Ablation Study (RQ2)}
Table~\ref{tab:ablation} summarizes two WIB-100-specific ablations: critic-penalty tuning and the replacement of theory-grounded connectors with generic prompts.

\begin{table*}[!t]
\caption{Ablation on critic falsification strength and theory-grounded connectors.}
\label{tab:ablation}
\centering
\footnotesize
\setlength{\tabcolsep}{4pt}
\renewcommand{\arraystretch}{1.14}
\begin{tabularx}{\textwidth}{@{} P{0.24\textwidth} C{0.08\textwidth} C{0.08\textwidth} C{0.08\textwidth} C{0.08\textwidth} X @{}}
\toprule[1.5pt]
\multicolumn{6}{@{}l}{\textbf{(a) Critic-penalty tuning}} \\
\midrule
\textbf{Penalty ($\gamma$)} & \textbf{MCC} & \textbf{F1$_{\mathrm{pos}}$} & \textbf{Rec.} & \textbf{Spec.} & \textbf{Behavior Profile} \\
\midrule
$\gamma=0$ (No falsification) & 0.463 & 76.1\% & 94.5\% & 46.0\% & Permissive regime with weak rejection \\
$\gamma=1.0$ (Moderate) & 0.619 & 82.0\% & 89.0\% & 72.0\% & Good recall--rejection trade-off \\
\rowcolor{tablegray}
$\gamma=2.0$ (Optimal) & \textbf{0.666} & \textbf{83.7\%} & \textbf{86.0\%} & \textbf{80.5\%} & Best balance on hard negatives \\
$\gamma=4.0$ (Over-penalized) & 0.582 & 77.0\% & 71.0\% & 86.5\% & Over-skeptical; suppresses true positives \\
\bottomrule[1.5pt]
\end{tabularx}

\vspace{0.85em}

\begin{tabularx}{\textwidth}{@{} P{0.24\textwidth} C{0.08\textwidth} C{0.08\textwidth} X @{}}
\toprule[1.5pt]
\multicolumn{4}{@{}l}{\textbf{(b) Theory-grounded connectors}} \\
\midrule
\textbf{Variant} & \textbf{MCC} & \textbf{Mac-F1} & \textbf{Interpretation} \\
\midrule
\rowcolor{tablegray}
\textbf{Full system (W{\"o}lfflin+IC)} & \textbf{0.666} & \textbf{83.2\%} & Formal and iconographic operators preserve discriminative structure \\
Generic prompts (no theory) & 0.501 & 73.0\% & Generic prompts smooth over diagnostic differences \\
\bottomrule[1.5pt]
\end{tabularx}
\end{table*}

The balanced WIB-100 benchmark makes the role of falsification especially transparent. With $\gamma=0$, the model recalls almost every positive (94.5\%) but rejects only 46.0\% of negatives, leading to 0.463 MCC. Increasing the critic penalty to $\gamma=2.0$ yields the best operating point: specificity climbs to 80.5\% while recall remains high at 86.0\%, producing the best overall F1 and MCC. Raising the penalty further to $\gamma=4.0$ yields an over-skeptical regime, where specificity rises again but recall collapses.

A related question is whether the theory-grounded operators can be replaced by generic prompts. Replacing the rigid W{\"o}lfflin/ICONCLASS operators with generic prompts causes a qualitatively different failure mode. The drop from 83.2\% to 73.0\% Macro-F1 and from 0.666 to 0.501 MCC shows that free-form language alone is too general: it can describe resemblance fluently, but it does not preserve enough formal or iconographic specificity to support reliable falsification.

\subsection{Leakage Control and Module Attribution}
\label{subsec:leakage_module}

\begin{table}[!t]
\caption{Leakage-control and module-level analyses on WIB-100.}
\label{tab:leakage_module}
\centering
\footnotesize
\setlength{\tabcolsep}{3pt}
\renewcommand{\arraystretch}{1.14}
\begin{tabularx}{\columnwidth}{@{} >{\hsize=1.4\hsize}L *{4}{>{\hsize=0.9\hsize}Z} @{}}
\toprule[1.5pt]
\multicolumn{5}{@{}l}{\textbf{(a) Biography leakage control}} \\
\midrule
\textbf{Condition} & \textbf{Mac-F1} & \textbf{MCC} & \textbf{Rec.} & \textbf{Spec.} \\
\midrule
\rowcolor{tablegray}
\textbf{Full biography} & \textbf{83.2\%} & \textbf{0.666} & \textbf{86.0\%} & \textbf{80.5\%} \\
Masked biography & 81.8\% & 0.637 & 83.2\% & 80.5\% \\
Text-only (no visual) & 71.2\% & 0.452 & 85.5\% & 58.0\% \\
No \Tool{BiographyReader} & 69.7\% & 0.395 & 68.5\% & 71.0\% \\
\midrule
\multicolumn{5}{@{}l}{\textbf{(b) Module-level architectural ablation}} \\
\midrule
\textbf{Variant} & \textbf{Mac-F1} & \textbf{MCC} & \textbf{Rec.} & \textbf{Spec.} \\
\midrule
\rowcolor{tablegray}
\textbf{Full \Agent} & \textbf{83.2\%} & \textbf{0.666} & \textbf{86.0\%} & \textbf{80.5\%} \\
w/o \Tool{TimelineGate} & 76.8\% & 0.549 & 86.0\% & 68.0\% \\
w/o \Tool{ConceptRetriever} & 79.0\% & 0.581 & 82.5\% & 75.5\% \\
w/o \Tool{StyleComparator} & 74.4\% & 0.493 & 80.0\% & 69.0\% \\
w/o \Tool{BiographyReader} & 69.7\% & 0.395 & 68.5\% & 71.0\% \\
w/o critic ($\gamma=0$) & 68.4\% & 0.463 & 94.5\% & 46.0\% \\
\bottomrule[1.5pt]
\end{tabularx}
\end{table}

Table~\ref{tab:leakage_module}(a) shows that masking explicit influence predicates causes only a modest drop from 83.2\% to 81.8\% Macro-F1 while leaving specificity unchanged at 80.5\%. This suggests that \Tool{BiographyReader} contributes more than verbatim label reading: the remaining pathway evidence still supports effective rejection of implausible links. The text-only setting further shows that biographies alone remain recall-heavy but substantially less specific than the full multimodal system.

Table~\ref{tab:leakage_module}(b) identifies how each module contributes. \Tool{TimelineGate} protects against temporal-impossible errors, \Tool{ConceptRetriever} and \Tool{StyleComparator} each sharpen specificity on hard visual confounders, and \Tool{BiographyReader} primarily supports recall by establishing plausible transmission pathways. The critic penalty remains important for suppressing the permissive YES bias that appears when falsification is removed.

\subsection{Negative-Tier Rejection Analysis (RQ3)}

\begin{figure*}[!t]
\centering
\includegraphics[width=0.98\textwidth]{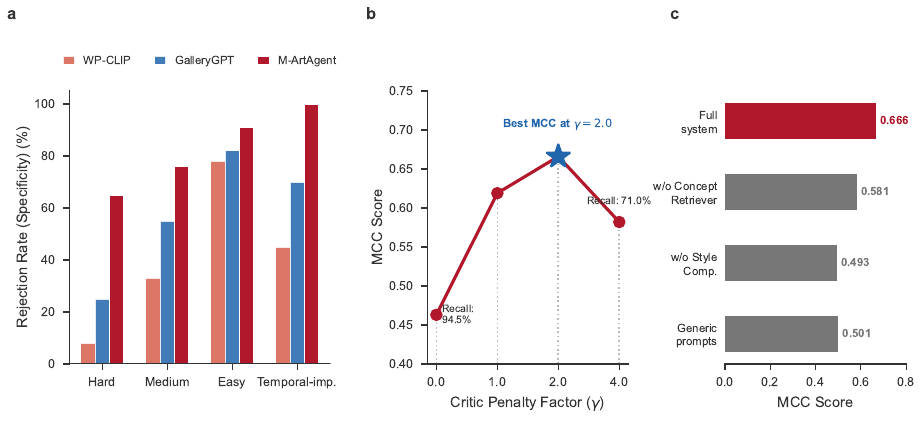}
\vspace{-0.5em}
\caption{Diagnostic analysis on the WIB-100 benchmark. (a) Stratified rejection rates for hard, medium, easy, and temporal-impossible negatives. (b) Effect of the critic penalty factor $\gamma$ on MCC. (c) Impact of replacing rigid formal connectors with generic prompts.}
\label{fig:error_robustness}
\end{figure*}

The decisive advantage of \Agent lies in rejecting the negatives that are hardest for similarity-based systems. On \emph{hard negatives}, namely pairs from the same era and movement but without direct influence, \Agent rejects about 65\% of pairs, compared with about 25\% for GalleryGPT and 8\% for WP-CLIP (Fig.~\ref{fig:error_robustness}\,(a)). On \emph{medium negatives}, which share an era but not a movement, the gap remains large: 76\% versus 55\% and 33\%, respectively.

Easy negatives are less informative because most systems can reject them once visual and historical differences become obvious. Even there, \Agent remains strongest at roughly 91\% rejection, compared with roughly 82\% for GalleryGPT and 78\% for WP-CLIP. The most striking result is the \emph{temporal-impossible} tier: the explicit timeline checker rejects 100\% of these cases, versus 70\% for GalleryGPT and 45\% for WP-CLIP. Weighted over the 1,000 negative pairs, these tier-level results exactly recover the overall 80.5\% specificity in Table~\ref{tab:main_results}.

\subsection{Operator Transparency Checks}
\label{subsec:transparency_checks}

\begin{table}[!t]
\caption{Additional transparency checks for the theory-grounded operators.}
\label{tab:transparency_checks}
\centering
\footnotesize
\setlength{\tabcolsep}{4pt}
\renewcommand{\arraystretch}{1.14}
\begin{tabularx}{\columnwidth}{@{} P{0.45\columnwidth} C{0.14\columnwidth} C{0.14\columnwidth} X @{}}
\toprule[1.5pt]
\multicolumn{4}{@{}l}{\textbf{(a) \Tool{ConceptRetriever} alignment (100 images)}} \\
\midrule
\textbf{Retrieval Level} & \textbf{Prec.} & \textbf{Rec.} & \textbf{F1} \\
\midrule
Exact match (Level 5+) & 68.4\% & 52.1\% & 59.1\% \\
Ancestor match (Level 3+) & 89.2\% & 81.5\% & 85.2\% \\
\bottomrule[1.5pt]
\end{tabularx}

\vspace{0.85em}

\begin{tabularx}{\columnwidth}{@{} P{0.45\columnwidth} C{0.14\columnwidth} C{0.14\columnwidth} X @{}}
\toprule[1.5pt]
\multicolumn{4}{@{}l}{\textbf{(b) W{\"o}lfflin manifold robustness}} \\
\midrule
\textbf{Configuration} & \textbf{Mac-F1} & \textbf{MCC} & \textbf{Impact} \\
\midrule
\rowcolor{tablegray}
\textbf{Standard full system} & \textbf{83.2\%} & \textbf{0.666} & Base \\
Axis-order perm. & 82.9\% & 0.660 & $-0.3$ pt \\
Prompt paraphrase & 83.5\% & 0.671 & $+0.3$ pt \\
Generic prompts & 73.0\% & 0.501 & Major drop \\
Alt. CLIP enc. & 79.4\% & 0.598 & Bounded \\
\bottomrule[1.5pt]
\end{tabularx}
\end{table}

Table~\ref{tab:transparency_checks}(a) clarifies the image-to-iconography interface. Exact leaf-code agreement is only moderate, but ancestor-aware matching is substantially stronger, which supports the use of topology-aware distances rather than exact-code equality alone. In other words, \Tool{ConceptRetriever} is more reliable as a semantic neighborhood operator than as a strict leaf-classifier.

Table~\ref{tab:transparency_checks}(b) shows that the W{\"o}lfflin manifold is not arbitrarily fragile: axis-order permutation causes only a minimal drop, and prompt paraphrases slightly improve performance. At the same time, the alternative-encoder result confirms that the manifold still depends on the underlying visual representation, so robustness should be interpreted as bounded rather than absolute.

\subsection{Case Studies (RQ5)}

\begin{figure*}[!t]
\centering
\includegraphics[width=0.98\textwidth]{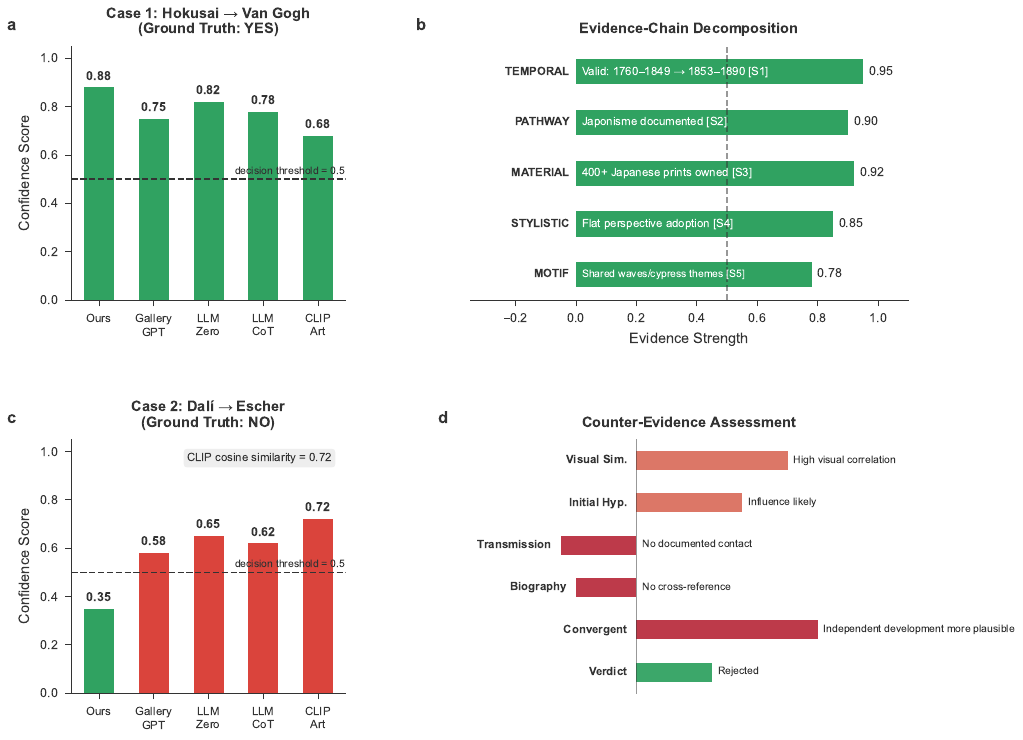}
\vspace{-0.5em}
\caption{Case studies. (a--b) True positive: Hokusai $\rightarrow$ Van Gogh. (c--d) True negative: Dal{\'i} $\rightarrow$ Escher, where high visual similarity is overturned by failed pathway evidence and counter-verification.}
\label{fig:case_study}
\end{figure*}
\subsubsection{Case 1: Hokusai $\rightarrow$ Van Gogh (True Positive)}

In the canonical Japonisme case of Hokusai $\rightarrow$ Van Gogh, whose ground truth is YES, \Agent outputs YES with confidence 0.88. The verdict is not driven by visual resemblance alone. \Tool{TimelineGate} establishes a feasible exposure window, \Tool{BiographyReader} retrieves strong pathway evidence centered on Japonisme and print collecting, and \Tool{VisualAnalyzer} together with \Tool{StyleComparator} identifies diagnostic formal markers that cohere with the retrieved context, such as planar compositional flattening. The evidence-chain panel in Fig.~\ref{fig:case_study}\,(b) shows that the temporal, pathway, material, stylistic, and motif signals all remain above the decision threshold.

\subsubsection{Case 2: Dal{\'i} $\rightarrow$ Escher (True Negative)}

Dal{\'i} $\rightarrow$ Escher illustrates a difficult true-negative case: the CLIP visual similarity is high (0.72), and several baseline systems therefore drift above the acceptance threshold. \Agent instead treats high similarity as a hypothesis to be falsified. The agent fails to recover robust transmission signals from biographies, such as documented contact or explicit cross-reference, and the critic elevates convergent development as the more plausible explanation. The counter-verification panel in Fig.~\ref{fig:case_study}\,(d) shows how the final confidence is driven down to 0.35, producing the correct NO verdict despite the deceptive visual match.

\subsection{Discussion}
\label{sec:discussion}

Three findings stand out from the evaluation. First, benchmark balance matters: once the positive prior is fixed at 50\%, trivial YES-heavy behavior is immediately exposed, and the evaluation rewards genuine discrimination rather than optimism. Second, stronger comparators matter: the added KG baselines improve over earlier metric-learning systems but remain well below \Agent, indicating that artist-level relational structure alone does not replace pair-specific evidence adjudication. Third, the main gain of \Agent is not in easy cases but in historically confounded ones. Timeline axioms, the theory-grounded operators, and critic-based falsification together are what turn resemblance into a historically plausible or implausible claim.

The leakage-control and transparency checks sharpen this methodological lesson. Masking explicit influence predicates only modestly reduces performance (1.4 points of Macro-F1), suggesting that \Tool{BiographyReader} does more than copy the answer string from the corpus. Likewise, the concept-alignment and manifold-robustness analyses show that the two theory-grounded operators are imperfect but operationally meaningful: replacing them with generic prompts reduces Macro-F1 by roughly 10 points, while axis-order and prompt-perturbation checks show bounded sensitivity. A purely visual similarity model can retrieve neighbors, and a knowledge-graph embedding can encode artist-level priors, but neither can decide whether a concrete pair reflects direct influence, shared conventions, or convergent evolution. \Agent works because it turns art-historical theory into first-class computational operators and then forces each hypothesis to survive counter-verification.

\subsubsection{Computational Cost and Scalability}

Because the pipeline mixes lightweight local computation with LLM-based reasoning, a layered complexity analysis is informative. Let $N_a$ denote the number of artists, $|\mathcal{A}_i|$ the portfolio size of artist $i$, $N=\sum_i|\mathcal{A}_i|$ the total number of artworks, $d$ the embedding dimension, and $M$ the maximum ReAct steps.

\textbf{Layer~2 (local computation).} Visual and textual encoding are single forward passes per artwork, costing $\mathcal{O}(N \cdot d)$ in total. The FAISS-HNSW index is built once in $\mathcal{O}(N\log N)$ time and queried in $\mathcal{O}(\log N)$ per artwork. Candidate generation therefore runs in $\mathcal{O}(N\log N)$ time and $\mathcal{O}(N\cdot d)$ space (for storing all embeddings).

\textbf{Layer~3 (LLM-based reasoning).} Each ReAct step ingests the accumulated evidence context and produces an action or conclusion. With a maximum of $M$ steps and a per-step context window of at most $L$ tokens, the worst-case token consumption per artist pair is $\mathcal{O}(M \cdot L)$. Under the current configuration ($M=8$, $L \approx 600$), this yields an upper bound of approximately 5,000 tokens per pair. For $P$ promoted candidate pairs, the total LLM token budget scales as $\mathcal{O}(P \cdot M \cdot L)$.

\textbf{End-to-end pipeline.} Without candidate filtering, all $\mathcal{O}(N_a^2)$ artist pairs would enter Layer~3, making LLM cost the dominant bottleneck. The Layer~2 candidate-generation module reduces this to $P \ll N_a^2$ high-likelihood pairs (approximately 200--500 in the WIB-100 setting) via FAISS retrieval and timeline gating, so the effective cost is $\mathcal{O}(N\log N + P \cdot M \cdot L)$. On the WIB-100 benchmark, evaluating all 2,000 labeled pairs with five-fold cross-validation consumes approximately 6--10M tokens in total, representing a nontrivial API cost that currently limits the system to research-scale deployment. Scaling to museum-scale collections (thousands of artists) would benefit from LLM distillation to smaller backbones, response caching across structurally similar pairs, or a tiered strategy that reserves full ReAct adjudication for borderline candidates.

\section{Conclusion}
\label{sec:conclusion}

This paper tackled a core problem in computational art history: discovering implicit art influence relationships that are visually plausible yet weakly documented. Three gaps were identified in existing approaches, namely historical constraint handling, falsification, and interpretability, and \Agent{} was proposed as a multimodal framework that reformulates influence detection as evidence-based adjudication rather than similarity ranking.

The system spans four layers, from data ingestion and dual-tower perception to ReAct-based reasoning and knowledge-graph materialization, mirroring art-historical practice while preserving an auditable evidence trail. On the balanced WIB-100 benchmark of 100 artists and 2,000 directed pairs, \Agent{} achieves 83.7\% positive-class F1, 0.666 MCC, and 0.910 ROC-AUC. The strongest gains come from rejecting hard negatives and all temporal-impossible negatives, and leakage-control, KG-baseline, and operator-transparency analyses indicate that these gains are not well explained by graph priors or explicit biography phrases alone.

That said, the present study is bounded by the scope of WIB-100 and by the availability of surviving documentation. Although the benchmark improves region and era coverage, \Agent{} is still evaluated on a 100-artist WikiArt-centered setting, so genuine but weakly documented influences may remain under-labeled. The pipeline also assumes access to both visual portfolios and textual biographies, leaving missing-modality settings under-explored, a limitation that echoes broader multimodal robustness concerns \cite{shi2024mora}. Moreover, the system's reliance on an LLM introduces cost, latency, and reproducibility variability, and the present experiments measure pair-level adjudication rather than end-to-end recall in a fully open-world setting.

Several promising extensions follow from these constraints. Extending the framework to non-Western traditions will require culturally grounded axioms and iconographic resources, not just geographic expansion. For example, applying the system to Chinese ink painting or Japanese \emph{ukiyo-e} would demand replacing the W{\"o}lfflinian formal axes with indigenous aesthetic frameworks, such as the Six Principles of Xie He or the Rinpa-lineage vocabulary, while adapting \Tool{ConceptRetriever} to East Asian iconographic taxonomies rather than the Eurocentric ICONCLASS hierarchy. More broadly, the four-phase adjudication protocol itself is domain-agnostic: adapting it to other domains such as music or literature would primarily require substituting the perceptual encoders (e.g., audio embeddings for music, language model embeddings for literary style) and the domain axioms, while the ReAct reasoning and adversarial falsification machinery would remain unchanged. Distinguishing finer-grained influence types, including compositional, thematic, and conceptual transmission, would sharpen the verdict semantics. Local or efficiently adapted multimodal backbones may improve reproducibility and reduce dependence on application programming interfaces (APIs); recent advances in parameter-efficient fine-tuning \cite{si2026tsd,si2025liera} and targeted knowledge updating \cite{shi2025dualedit} suggest concrete routes. Finally, human--AI interfaces that let art historians inspect evidence, guide tool use, and override verdicts would bring the system closer to practical scholarship. However, its outputs should remain decision support: confidence scores are not certainty, and expert validation remains essential before publication.

In summary, \Agent{} demonstrates that implicit art influence analysis benefits from treating the task as historically constrained adjudication rather than pattern matching alone. By combining multimodal perception, iterative reasoning, domain constraints, and adversarial falsification, the framework delivers strong gains over representative baselines while remaining interpretable. This work aims to encourage closer collaboration between computational methods and art-historical scholarship.

\bibliographystyle{IEEEtran}
\bibliography{refs}

\begin{biography}[{\includegraphics[width=1in,height=1.25in,clip]{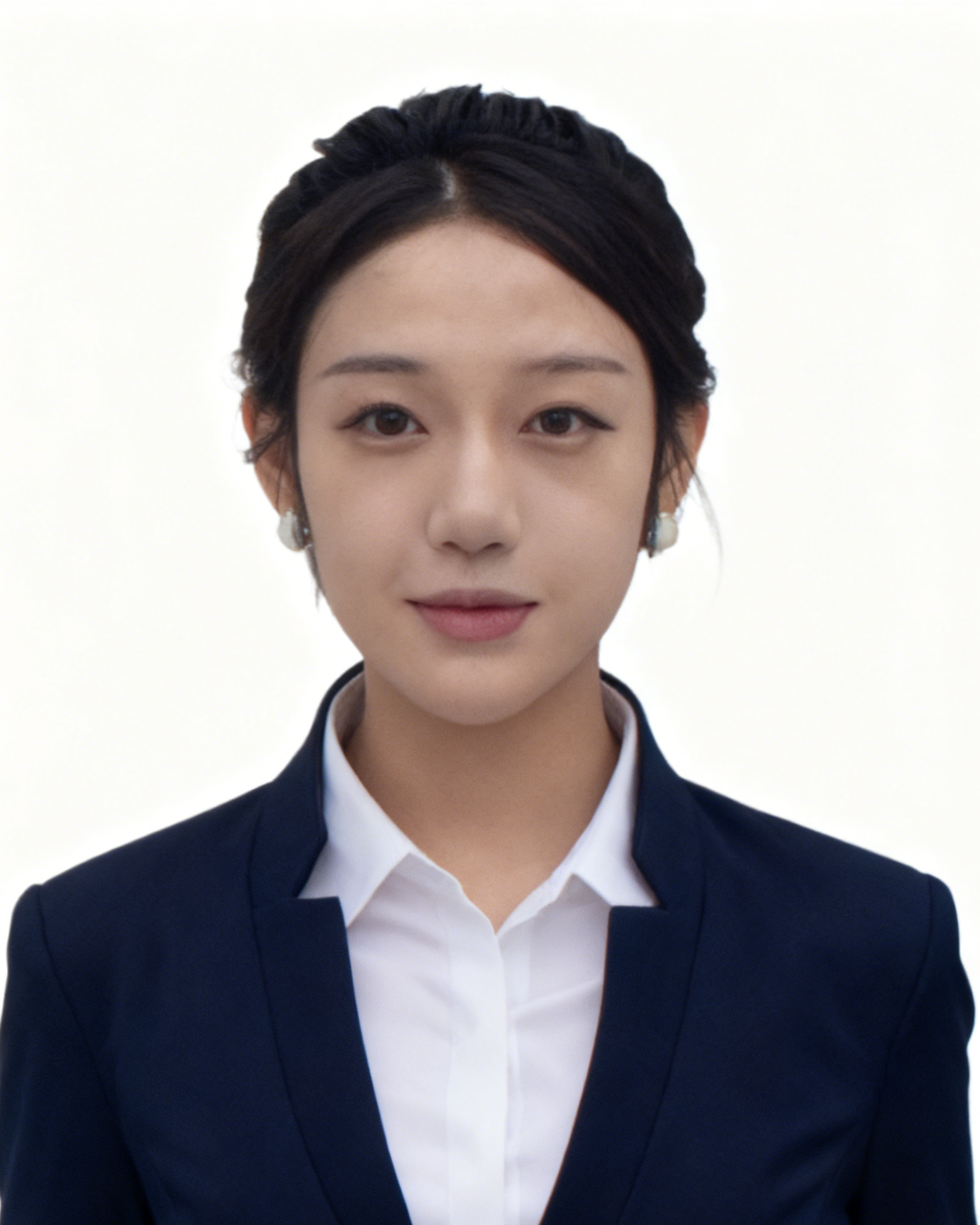}}]{Hanyi Liu}
Hanyi Liu received the B.S. degree from Southeast University, Nanjing, China, and the M.A. degree from the Royal College of Art, London, U.K. She is currently a researcher with China Electronics Technology Group Co., Ltd., Beijing 100043, China. She has more than two years of experience in computer product research and development within technical engineering teams. Her interdisciplinary background spans computer science and art studies, and her research interests include multimodal artificial intelligence, computational analysis of visual culture, and artificial-intelligence-driven methods for cultural heritage and art research.
\end{biography}

\begin{biography}[{\includegraphics[width=1in,height=1.25in,clip]{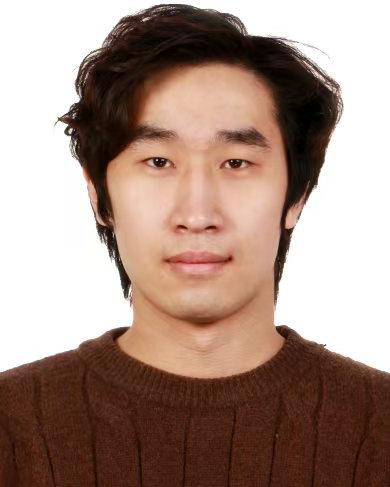}}]{Zhonghao Jiu}
Zhonghao Jiu received the B.S. degree in information science and engineering from the Wu Jianxiong Honors College, Southeast University, Nanjing, China. He is currently pursuing the Ph.D. degree as a direct-entry doctoral student with the School of Information Science and Engineering, Southeast University, Nanjing, China. His research focuses on large language models and applications of knowledge graphs.
\end{biography}

\begin{biography}[{\includegraphics[width=1in,height=1.25in,clip]{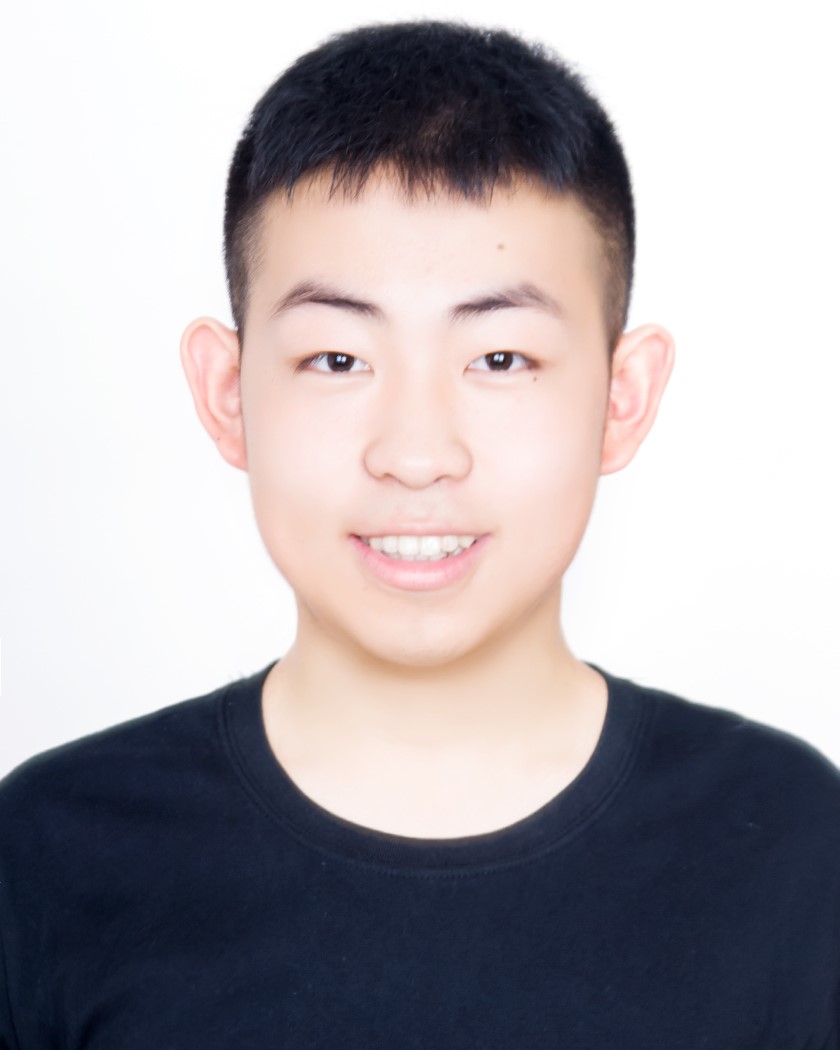}}]{Minghao Wang}
Minghao Wang is with the Department of Chemical and Biological Engineering, Hong Kong University of Science and Technology, Hong Kong SAR, China. His research interests include large language models and recommender systems.
\end{biography}

\begin{biography}[{\includegraphics[width=1in,height=1.25in,clip]{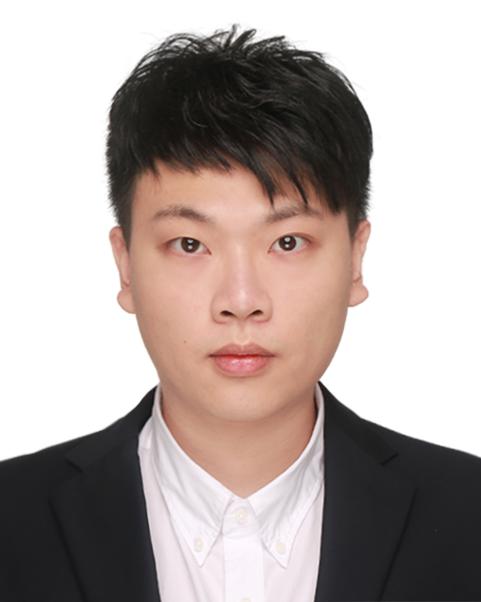}}]{Yuhang Xie}
Yuhang Xie received the B.S. degree in software engineering from Peking University, Beijing, China. He is currently pursuing the M.S. degree in computer science with the University of California San Diego, La Jolla, CA, USA. His research interests include large language models, distributed systems, and computer security.
\end{biography}

\begin{biography}[{\includegraphics[width=1in,height=1.25in,clip]{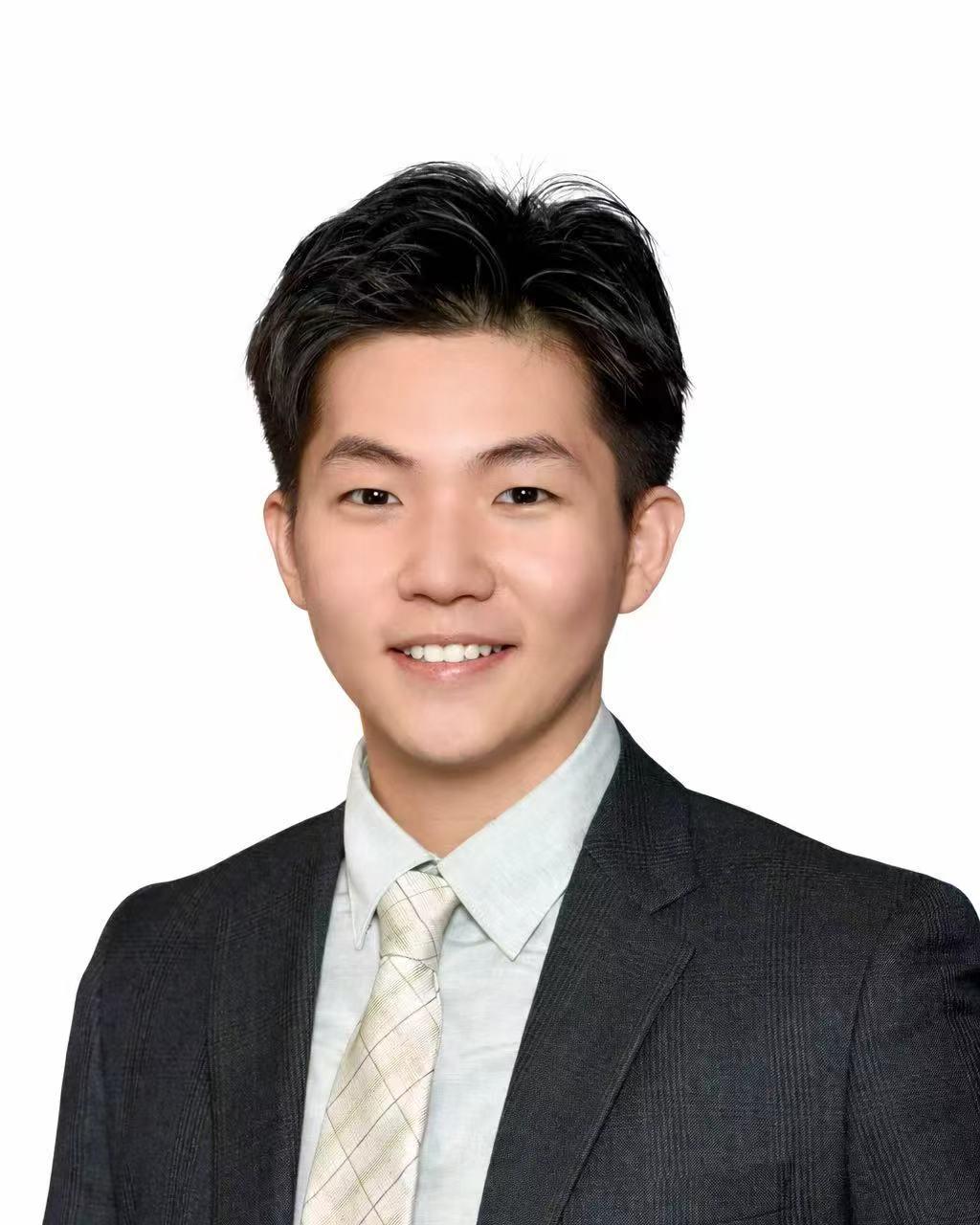}}]{Heran Yang}
Heran Yang received the B.S. degree in computer science and business from Northeastern University, Boston, MA, USA. His research interests include artificial intelligence for healthcare systems.
\end{biography}

\EOD

\end{document}